\documentclass[letterpaper]{article} 
\usepackage{aaai24}  
\usepackage{times}  
\usepackage{helvet}  
\usepackage{courier}  
\usepackage[hyphens]{url}  
\usepackage{graphicx} 
\usepackage{natbib}  
\usepackage{caption} 
\usepackage{algorithm}
\usepackage{amsmath}
\usepackage{bm}
\usepackage{amsthm}

\usepackage{bibentry}
\usepackage{booktabs}

\usepackage{float}
\usepackage{newfloat}
\usepackage{listings}
\DeclareCaptionStyle{ruled}{labelfont=normalfont,labelsep=colon,strut=off} 
\floatstyle{ruled}
\newfloat{listing}{tb}{lst}{}
\floatname{listing}{Listing}
\usepackage{algpseudocode}
\usepackage{multicol}
\usepackage{amsfonts}
\usepackage{xcolor}
\usepackage{booktabs} 
\usepackage{bbding}
\usepackage{float}
\usepackage{array}

\newcommand{\name}{\textit{Chat2Layout }}

\definecolor{Maroon}{RGB}{128, 0, 0}
\definecolor{NavyBlue}{RGB}{0, 0, 128}
\definecolor{OliveGreen}{RGB}{107, 142, 35}
\definecolor{Thistle}{RGB}{216, 191, 216}
\definecolor{Purple}{RGB}{128, 0, 128}

\title{\textit{Chat2Layout:} Interactive 3D Furniture Layout with a Multimodal LLM}
\author {
    Can Wang\textsuperscript{\rm1},
    Hongliang Zhong\textsuperscript{\rm1},
    Menglei Chai\textsuperscript{\rm2},
    Mingming He\textsuperscript{\rm3},
    Dongdong Chen\textsuperscript{\rm4},
    Jing Liao\textsuperscript{\rm1}\thanks{Corresponding author},
}
\affiliations {
    \textsuperscript{\rm 1}
    City University of Hong Kong
    \textsuperscript{\rm 2}Google AR Perception
    \textsuperscript{\rm 3}Netflix Eyeline Studios  
    \textsuperscript{\rm 4}Microsoft Cloud AI

    \{cwang355-c, hlzhong2-c\}@my.cityu.edu.hk,
    cmlatsim@gmail.com,
    mheah@connect.ust.hk,\\
    cddlyf@gmail.com,
    jingliao@cityu.edu.hk
}

\begin{document}

\maketitle

\begin{abstract}
{Automatic furniture layout is long desired for convenient interior design.
Leveraging the remarkable visual reasoning capabilities of multimodal large language models (MLLMs), recent methods address layout generation in a static manner, lacking the feedback-driven refinement essential for interactive user engagement.
We introduce \textit{Chat2Layout}, a novel interactive furniture layout generation system that extends the functionality of MLLMs into the realm of interactive layout design.
To achieve this, we establish a unified vision-question paradigm for in-context learning, enabling seamless communication with MLLMs to steer their behavior without altering model weights. Within this framework, we present a novel training-free visual prompting mechanism. This involves a visual-text prompting technique that assist MLLMs in reasoning about plausible layout plans, followed by an Offline-to-Online search (O2O-Search) method, which automatically identifies the minimal set of informative references to provide exemplars for visual-text prompting. By employing an agent system with MLLMs as the core controller, we enable bidirectional interaction. The agent not only comprehends the 3D environment and user requirements through linguistic and visual perception but also plans tasks and reasons about actions to generate and arrange furniture within the virtual space. Furthermore, the agent iteratively updates based on visual feedback from execution results. Experimental results demonstrate that our approach facilitates language-interactive generation and arrangement for diverse and complex 3D furniture.}
\end{abstract}

\section{Introduction}

\begin{figure*}[t]
\centering
\setlength{\tabcolsep}{0\linewidth}
\includegraphics[width=1.0\textwidth]{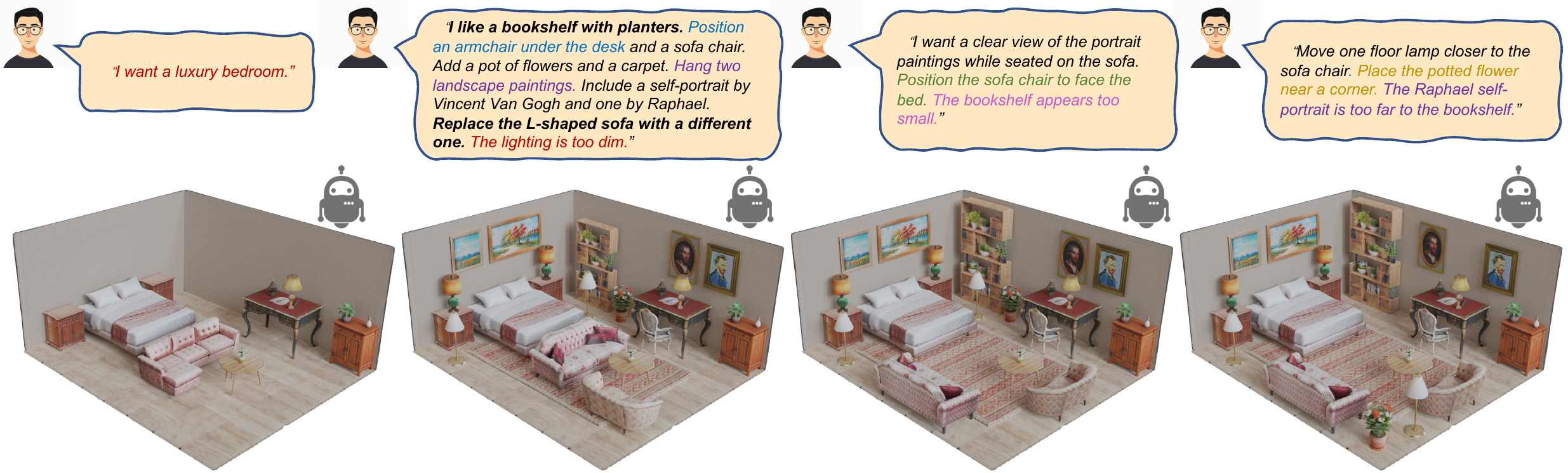}
\caption{\textit{Chat2Layout} features a multimodal LLM agent designed to facilitate natural language interaction between users and 3D indoor environments. Users can provide a wide range of instructions, from abstract requests (texts in {\color{Maroon}red}) to specific commands (texts in {\color{NavyBlue}blue}), whether in isolation (the first chat) or as part of a continuous dialogue (subsequent interactions). The agent interprets these instructions to autonomously executes corresponding operations within the 3D environment. This enables a multi-turn conversation process, empowering users to provide feedback and engage dynamically with the environment. \textit{Chat2Layout} supports a variety of applications, including object removal or addition (texts in \textbf{bold}), rotation (texts in {\color{OliveGreen}green}), scaling (texts in {\color{Thistle}pink}), and re-arrangement (texts in {\color{brown}yellow}). Additionally, it enables 3D layout generation that involves walls (texts in {\color{Purple}purple}).}
\label{fig:teaser}
\end{figure*}

3D furniture layout generation plays a crucial role in various applications such as interior design, game development, and virtual reality. Traditional methods for furniture layout planning often formulate the task as a constrained optimization problem~\cite{qi2018human,ma2016action,chang2015text,chang2014learning,yeh2012synthesizing,fisher2012example,luo2020end,wang2019planit,ma2018language,fu2017adaptive}. However, these approaches typically require rules for scene formation and graph annotations, necessitating the expertise of professional artists. This makes these methods less accessible to non-professionals and inflexible to dynamic environments where user preferences and spatial constraints may vary. 

Recent learning-based approaches address these limitations by employing
neural networks \cite{zhang2020deep,wang2018deep,ritchie2019fast,zhou2019scenegraphnet, li2019grains, dhamo2021graph,paschalidou2021atiss,wang2021sceneformer,zhai2024commonscenes,tang2023diffuscene,lin2023instructscene}, 
to automate the object selection and placement.
Learning from large datasets, these generative methods can produce diverse plausible layouts \cite{deitke2023objaverse,fu20213df,fu20213d}. 
However, despite the promise, they face difficulties in accommodating to objects that are not presented in the training set, limiting their adaptability and versatility.

The rapid advancement of large language models (LLMs) has opened up new possibilities for enhancing user interactions in furniture layout generation. Researchers leverage LLMs to reason about layout plans based on text descriptions of furniture items, including their positions and dimensions \cite{wen2023anyhome,feng2024layoutgpt}. These approaches enable users to express their desired living spaces in natural language, eliminating the need for extensive training on large datasets.
However, these methods suffer from notable limitations. While some of these methods~\cite{wen2023anyhome} utilize visual information to refine placement, they still primarily rely on textual input without fully incorporating visual perception. As a result, they often produce plausible yet impractical layouts that fail to align with user expectations or spatial constraints.
Furthermore, the absence of agent memory and feedback mechanisms prevents multi-turn conversations, hindering users' ability to interact with the process to iteratively refine the generated layout plan.

Recent developments have highlighted two significant trends in the field: 
First, Multimodal LLMs (MLLMs) \cite{openai2023vision,zhu2023minigpt,li2024llava,chu2024mobilevlm,zhan2024anygpt,li2024mini,dong2024internlm,chen2024far} have attracted huge attention for their impressive capabilities in various vision-language tasks by introducing visual perceptions into LLMs. 
Second, LLM agents \cite{belzner2023large,zhao2024expel,wu2023autogen,chan2023chateval,deng2024mind2web} emerge as powerful problem-solvers for handling complex tasks through feedback mechanisms and iterative refinement.
Inspired by these advancements, we aim to develop a MLLM agent specifically designed for generating furniture layouts.

In this paper, we introduce \textit{Chat2Layout}, a language-interactive method for generating furniture layouts that employs a MLLM as its agent. To begin, we establish a unified
vision-question paradigm for in-context learning, which standardizes communication with the MLLM to guide its reasoning utilizing both textual and visual information without the need to update model weights.
Within this framework, we present a novel training-free visual prompting mechanism composed of two key components:
1) We develop a visual-text prompting technique that assists the MLLM in tackling specific layout tasks;
2) We propose an Offline-to-Online search (O2O-Search) method to automatically identify the minimal support set from an example database, facilitating efficient in-context learning from a limited number of references.

Building upon these techniques, we develop a MLLM agent system that automatically perceives 3D indoor environments through both linguistic and visual modalities. The agent is capable of understanding user requirements, planning tasks, reasoning about the necessary actions to generate and arrange furniture within a virtual environment, and learning from the visual feedback of its executed results.
By leveraging our sophisticated unified vision-question paradigm, which incorporates both a visual-text prompting technique and an O2O-Search method, in conjunction with the MLLM agent system, our \textit{Chat2Layout} enables the execution of diverse, complex, and language-interactive 3D furniture generation and placement. This enables new user experiences that are previously unsupported, setting a new milestone in the field of furniture layout generation.

Our main contributions can be summarized as follows:
\begin{itemize}
\item A MLLM agent system that enables language-interactive 3D furniture generation and layout. This system supports a multi-turn conversations, allowing users to interact dynamically with the 3D environment and iteratively refine the layouts.
\item A unified vision-question paradigm for MLLM to effectively respond to a wide range of tasks. This paradigm incorporates a visual-text prompting technique that facilitates grid-based placement for layout generation, and a O2O-Search method that significantly boosts MLLM's reasoning capabilities in generating furniture plans.
\item Comprehensive support for various furniture layout applications, including layout completion, rearrangement, open-set placement, and multi-conventional interaction, as shown in Figure \ref{fig:teaser}.
\end{itemize}

\section{Related Work}

\noindent{\textbf{Furniture Layout Generation.}}
Traditional optimization-based layout generation~\cite{qi2018human,ma2016action,chang2015text,chang2014learning,yeh2012synthesizing,fisher2012example,luo2020end,wang2019planit,ma2018language,fu2017adaptive} relies on prior knowledge of reasonable configurations, such as procedural modeling or pre-defined scene graphs. Such priors require professional expertise, and are less flexible in dynamic environments.

Generative methods address these challenges with CNN \cite{zhang2020deep,wang2018deep,ritchie2019fast,zhou2019scenegraphnet}, 
VAE \cite{li2019grains},
GCN \cite{dhamo2021graph},
transformer \cite{paschalidou2021atiss,wang2021sceneformer}, or diffusion \cite{zhai2024commonscenes,tang2023diffuscene,lin2023instructscene} architectures trained on large-scale datasets \cite{deitke2023objaverse,fu20213df,fu20213d}. 
\textit{InstructScene} \cite{lin2023instructscene} handles concrete layout instructions by training a generative model on a scene-instruction dataset, incorporating a semantic graph prior and a diffusion decoder.
However, these methods struggle with placing objects not included in the training data. In contrast, \name supports open-set furniture placement with required furniture pieces automatically generated.

LLMs offer a new sight of text-based layout reasoning, bypassing the dataset limitations.
\textit{LayoutGPT} \cite{feng2024layoutgpt} enables LLMs to deliberate over layout plans with bounding boxes and orientation of furniture items.
\textit{AnyHome} \cite{wen2023anyhome} enhances \textit{LayoutGPT} with manually defined placement rules via text instructions.
While \textit{AnyHome} utilizes visual information, it primarily relies on textual input for placement and only uses visual cues for minor adjustments. Without proper visual understanding, these methods can produce plausible but impractical results. Additionally, the absence of agent memory and feedback mechanisms prevents multi-turn conversations, hindering users from interacting with the process to iteratively refine the generated layout plan.

\noindent{\textbf{LLMs as Agents.}}
With their potential to be powerful general problem solvers, LLMs have been used as the core controller for agents \cite{belzner2023large,zhao2024expel,wu2023autogen,chan2023chateval,deng2024mind2web,AutoGPT,gpt-engineer,BabyAGI}. 
Recently, MLLMs \cite{openai2023vision,zhu2023minigpt,li2024llava,chu2024mobilevlm,zhan2024anygpt,li2024mini,dong2024internlm,chen2024far} demonstrate even more impressive performance 
by incorporating visual perception.
Yet, MLLMs as agents has not been fully explored. We are the first to design such an agent system for furniture layout generation, facilitating continuous layout arrangements.

\noindent{\textbf{In-context Learning.}}
ICL integrates task demonstrations into prompts to enhance the reasoning ability of LLMs,
such as Chain-of-Thought (CoT) \cite{wang2022self} and automatic prompt design \cite{zhang2022automatic, shin2020autoprompt, zhou2022large, lester2021power}. Recently, visual prompting has been used in vision tasks, such as overlaying masks with a red circle \cite{shtedritski2023does}, highlighting regions \cite{yang2024fine}, using multiple circles with arrows \cite{yang2023dawn}, and labeling objects with alphabet numbers \cite{yang2023set}. 
Other techniques select representative exemplars from a candidate set for prompting. These reference exemplars are defined as support set.
Offline methods calculate pairwise similarities among the dataset for selecting the top-$k$ exemplars \cite{su2022selective},
while online methods calculate the similarity between the test prompt and candidates to select exemplars \cite{alayrac2022flamingo,yang2022empirical,feng2024layoutgpt}.  
These methods have not been fully explored for visual tasks with MLLMs like ours.

\noindent{\textbf{Text-to-3D Generation.}}
For layout visualization, 
current methods often pre-select or retrieve furniture objects from a 3D dataset,
limiting the ability to meet diverse user expectations.
Instead, our method uses text-to-3D generation to create furniture items.
Text-to-3D generations \cite{Meshy,Luma,Sudo} create 3D content like mesh \cite{ma2023x,mohammad2022clip}, neural radiance field \cite{wang2023nerf,lin2023magic3d,raj2023dreambooth3d,wang2024prolificdreamer,hong20243dtopia}, or gaussian splatting \cite{tang2024lgm,yi2023gaussiandreamer,tang2023dreamgaussian} from text prompts in an optimization \cite{poole2022dreamfusion,haque2023instruct} or generative manner \cite{wang2022clip,metzer2023latent}.
Compared to generative methods, recent optimization-based approaches offer broader diversity with fast-optimization processes \cite{tang2024lgm, li2023instant3d}. 
In this work, we adopt Tripo3D \cite{tripo3d}, which is capable of producing high-quality 3D furniture swiftly.

\section{Overview of Our Agent}

\begin{figure*}[t]
\centering 
\includegraphics[width=0.98\textwidth]{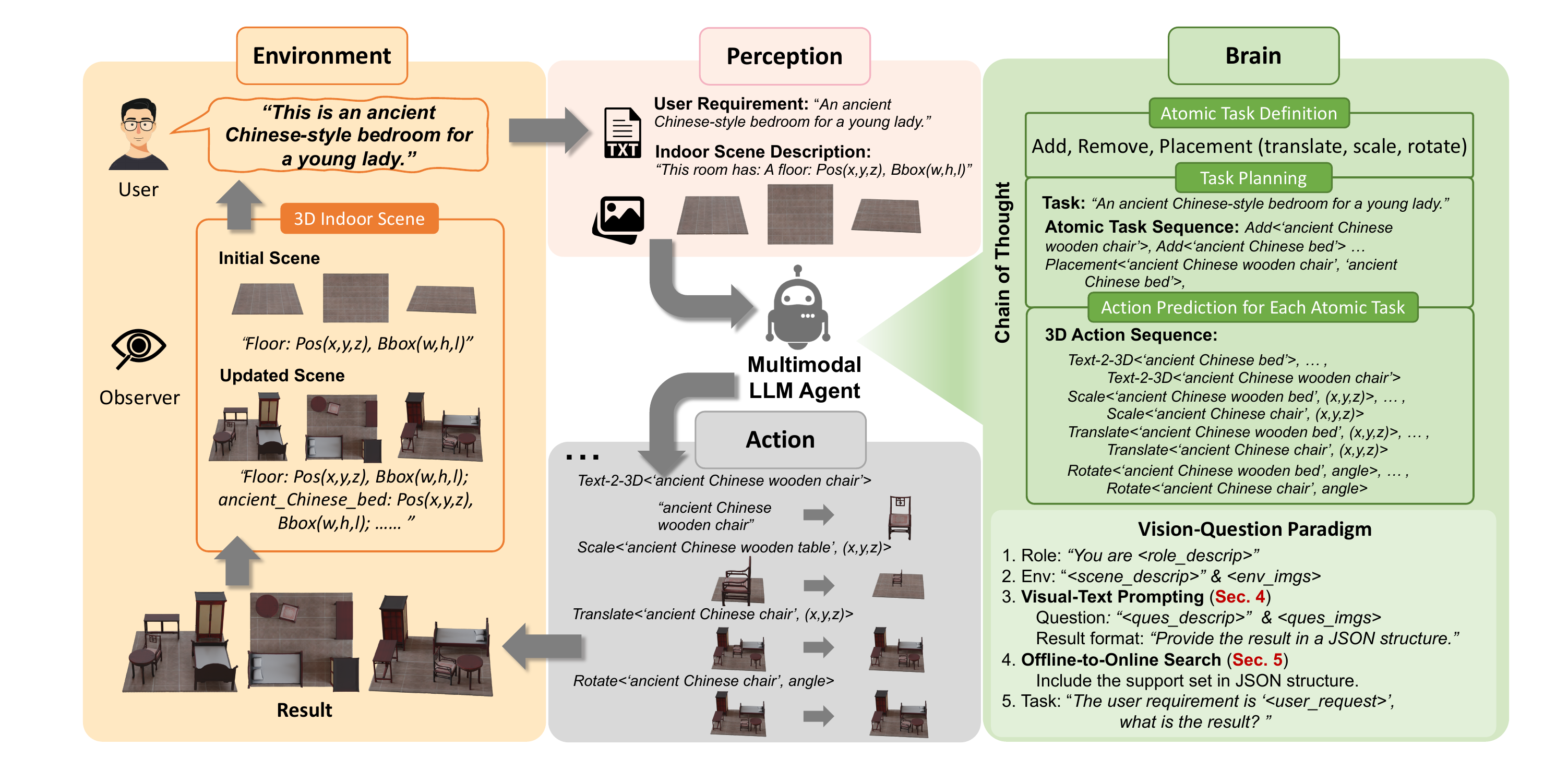}
\caption{\textbf{The framework of \textit{Chat2Layout}.} 
Our agent begins by perceiving the user's requirements, along with the indoor scene descriptions and visual captures provided by the observer from the environment. Within the brain module, the agent translating requirements into a task and decomposing it into a sequence of atomic tasks. 
It then reasons about the necessary 3D actions for each atomic task, generating a sequence of 3D actions that are subsequently executed in the action module, resulting in modifications to the environment.
Users can then observe the updated 3D indoor scene and provide new requirements.
Throughout this process, all interactions with the MLLM adhere to the vision-question paradigm for in-context learning, which adopts visual-text prompting to assist the agent in reasoning and decision-making and Offline-to-Online Search to provide the support set as references for prompting.
}
\label{fig:framework}
\end{figure*}

\subsection{Classical Agent System}

An agent system typically comprises four main components \cite{xi2023rise}:
\textbf{Environment:} The context within which the agent operates, such as web engines or mobile apps, defining the state space of the agent. The agent's actions can modify the environment, thereby influencing its own decision-making processes.
\textbf{Perception:} The agent's sensory component, collecting information from the environment through various modalities like vision and audio. The perceived input is transformed into neural signals and transmitted to the brain for further processing.
\textbf{Brain:} The central processing unit of the agent, responsible for storing knowledge and memories, as well as performing essential functions such as information processing and decision-making. It enables the agent to reason and plan, handle unforeseen tasks, and exhibit intelligent behaviors.
\textbf{Action:} The component that receives and executes action sequences from the brain, once the decisions are made, allowing the agent to interact with the environment.

\subsection{MLLM-Based Layout Agent}

\textit{Chat2Layout} draws inspiration from existing research on LLMs as agents, but it distinguishes itself by uniquely integrating an additional visual modality.
To create such an agent system specialized for interactive layout generation, we implement specific modifications to the agent framework (Figure \ref{fig:framework}):
\textbf{Environment:} Our environment includes a user and an observer. The user provides text-based requirements, while the observer monitors the 3D user interface to summarize visual elements and furniture attributes like positions and dimensions. The observer also engages in self-reflection, prompting the agent to review initial layouts and request corrections.
\textbf{Perception:} The agent perceives both textual and visual information from the environment, encompassing user requirements, furniture attributes, and visual room scenes.
\textbf{Brain:} Based on the perceived information, the agent reasons about the necessary 3D actions for layout plan generation.
\textbf{Action:} Once the 3D action list is determined, these actions are executed within the 3D visualization engine, bringing the layout plan to life.

\subsection{Brain}

Our core techniques are centered around the Brain module, the central decision-making component of our agent. Its input includes a specific user task and scene information perceived by the Perception module, while its output is a sequence of 3D actions that can be executed by the Action module. Every interaction with the MLLM adheres to our vision-question paradigm.
The Brain first decomposes each task into a series of atomic tasks. Then, for each atomic task, it performs visual-text prompting to formulate prompts incorporating references generated by O2O Search for engaging with the MLLM. Subsequently, it receives 3D actions, which are then passed on to the Action module for execution.

We give definitions of the most important terms of Brain below, and will describe its key technical components in following sections.

\noindent{\textbf{Task.}} 
A task is defined by user requirements described in text. \name supports open-vocabulary and unrestricted user descriptions, whether abstract or concrete, complex or simple, mixed or focused. This flexibility leverages the commonsense knowledge inherent in LLMs to guide and control the design generation process.

\noindent{\textbf{Atomic Task.}} 
An atomic task is the most basic unit of a task that cannot be further decomposed. We identify three specific types of atomic tasks: \textit{Add\textless text\textgreater }, \textit{Remove\textless text\textgreater }, and \textit{Placement(translate, scale, rotate)\textless text\textgreater }.

\noindent{\textbf{3D Action.}} 
We define five fundamental actions within the 3D engine to manipulate objects:
\begin{itemize}
    \item `\textit{Add\textless objkey, text\textgreater }': Invokes the text-to-3D API `\textit{Text-2-3D\textless text\textgreater }' to generate a textured mesh from a text description, then assigns an object key `\textit{objkey}' to this mesh.
    \item \textit{Remove\textless objkey\textgreater }: Removes the object identified by `\textit{objkey}' from the scene.
    \item \textit{Translate\textless objkey, (x,y,z)\textgreater }: Moves an object to a new position specified by the coordinates $(x,y,z)$.
    \item \textit{Scale\textless objkey, (x,y,z)\textgreater } - Adjusts the size of an object according to the scale factors.
    \item \textit{Rotate\textless objkey, angle\textgreater } - Rotates an object around its center by a specified angle.
\end{itemize}

\begin{figure}[t]
\centering
\includegraphics[width=0.98\linewidth]{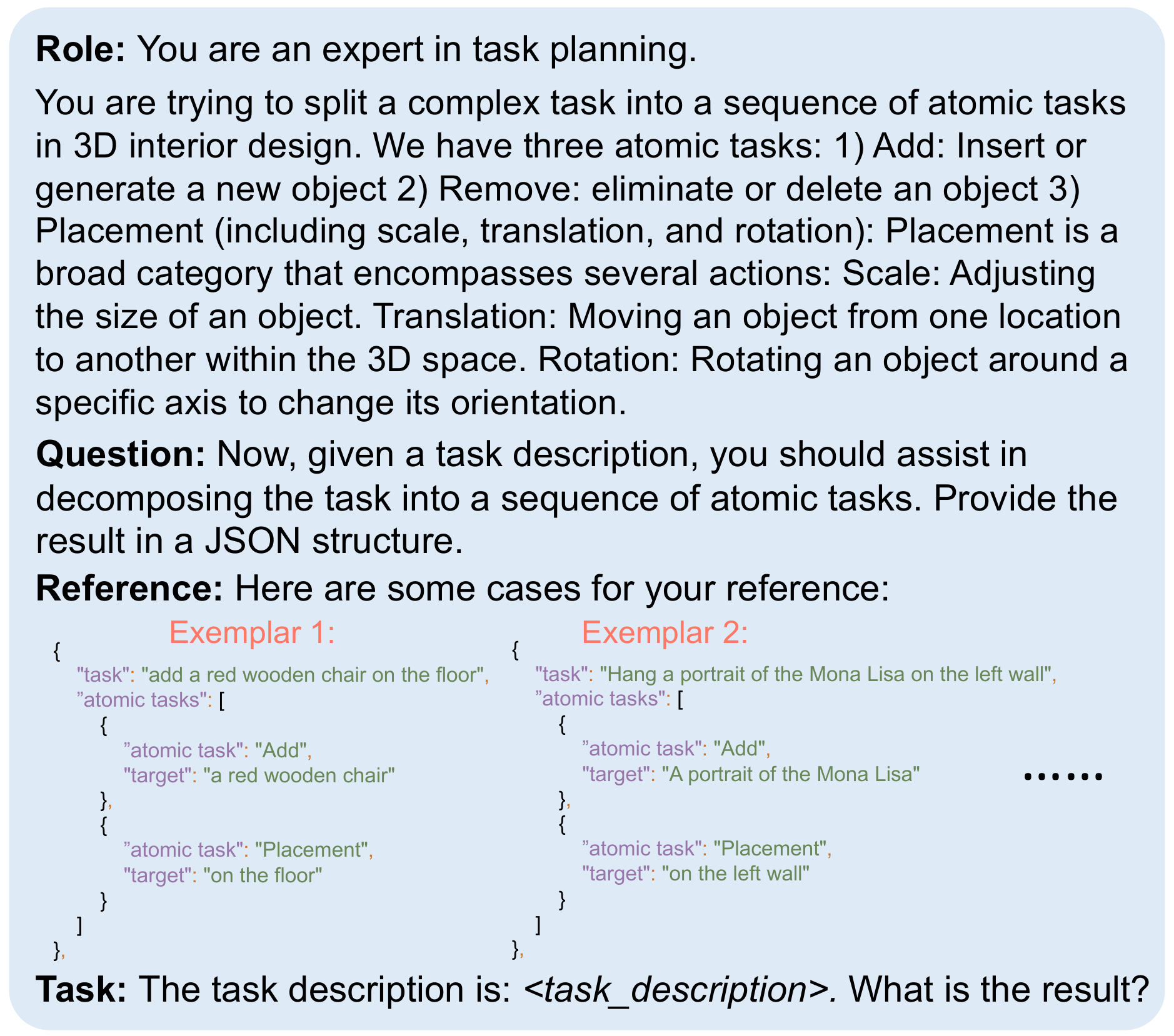}
\caption{\textbf{Vision-question paradigm for task decomposition.} We decompose a task into a sequence of atomic tasks using vision-question paradigm.}
\label{fig:method_taskdecom}
\end{figure}

\section{Vision-Question Paradigm}
\label{subsec:vqp}

We observe a similar prompt pattern in each interaction with the MLLM. Therefore, we adopt a unified vision-question paradigm as the template for every interaction, as depicted in Figure \ref{fig:framework}.

In this paradigm, we first assign the MLLM a \textbf{Role} such as ``You are an expert in 3D interior design'', which has proven crucial in enhancing the LLMs's performance. For specific tasks involving mathematical reasoning, such as object scale prediction and 3D placement, we add addition constraints as ``You are an expert in 3D interior design with a strong math background''.

We incorporate \textbf{Environment} information, includes textual descriptions like object names (represented by `\textit{objkey}'), positions, and dimensions of visualized objects, along with multi-view visual captures of the user interface.

We introduce a visual-text prompting method that formulating each specific \textbf{Question} with visual and textual information to aid the MLLM in reasoning about plausible layout plans.

Additionally, to improve MLLM's accuracy through in-context learning (ICL) from a few contextual examples, we propose an O2O-Search method for selecting the support set as \textbf{Reference}. This approach automatically identifies the minimal informative support set required for effective in-context learning.

By combining visual and textual information, our unified vision-question paradigm enables more accurate interpretations and responses for each \textbf{Task}, as visual cues help disambiguate textual input, leading to improved decision-making.

Exemplar prompts following our vision-question paradigm can be found in our supplementary material.

\paragraph{Task Decomposition}
Given a user task, before applying visual-text prompting, we need to first decompose it into a sequence of atomic tasks. Figure \ref{fig:method_taskdecom} illustrates the vision-question paradigm for task decomposition, including the assigned role and the support set provided by O2O-Search. The support set is presented in JSON format, suitable for representing structured data. Note that, as a special case of vision-question paradigm, environment and visual-text prompting are not included in task decomposition as it is a text-only problem. For other interactions discussed in Section \ref{subsec:vtp}, we adhere to the complete vision-question paradigm.

\begin{figure}[t]
\centering
\includegraphics[width=0.98\linewidth]{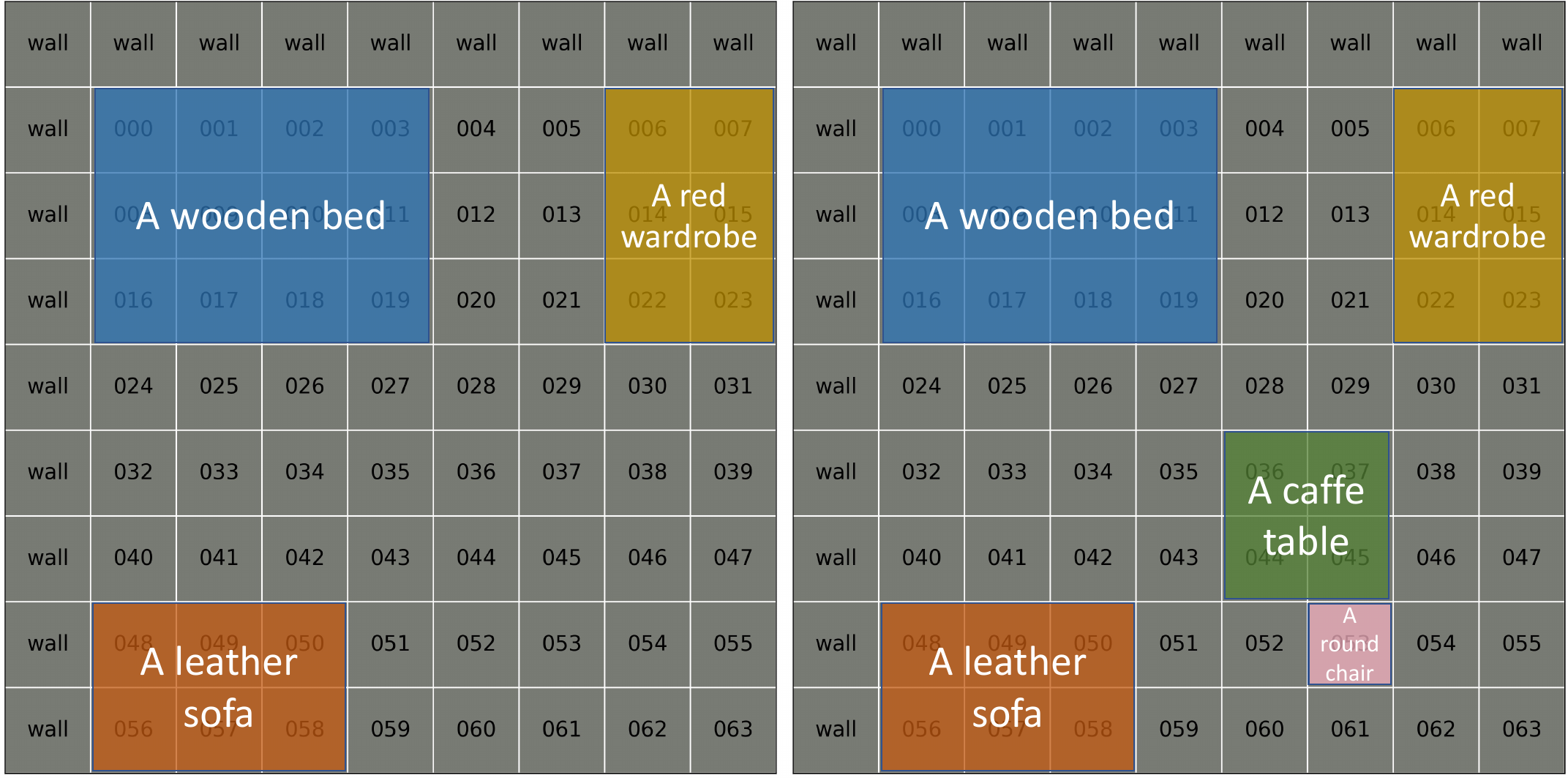}
\begin{small}
\begin{tabular}{c}
(a) Local by local placement
\end{tabular}
\end{small}
\includegraphics[width=0.98\linewidth]{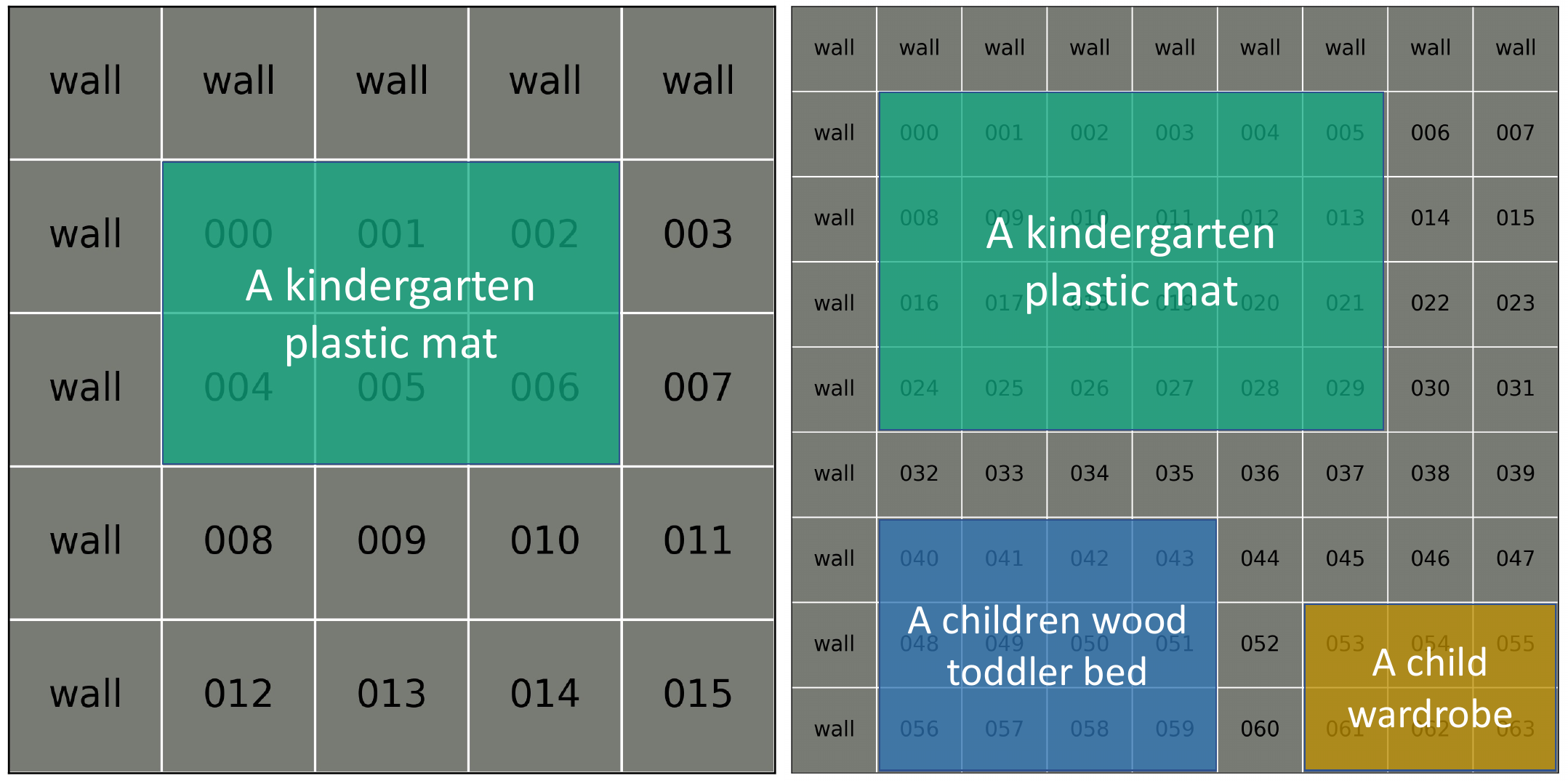}
\begin{small}
\begin{tabular}{c}
(b) Scale-adaptive grid placement
\end{tabular}
\end{small}
\caption{\textbf{Grid coverage algorithm for floor placement.} We conduct a local by local placement for a large set of furniture, and we propose a scale-adaptive grid placement for large-size furniture items. 
}
\label{fig:method_grd}
\end{figure}

\section{Visual-Text Prompting}
\label{subsec:vtp}

\subsection{Object Initialization}

We generate furniture items using `\textit{Text-2-3D\textless text\textgreater }', which come with arbitrary poses and sizes, which are not ideal for placement. Therefore, before incorporating them into a layout, we initialize them with properly aligned poses and scales using visual-text prompting.
In these paradigms, we provide environment information, including visual captures of the scene and object bounding boxes, along with references from our O2O-Search.

\noindent{\textbf{Pose Alignment.}}
First, we calculate the Axis-Aligned Bounding Box (AABB) and the Optimal/Oriented Bounding Box (OBB) \cite{fabri2009cgal}. 
Then, we align each object's pose by rotating it from its OBB to match its AABB.

\noindent{\textbf{Scale Prediction.}} 
We formulate a prompt to predict the appropriate scaling factor for each object. 
After adjusting the furniture items to their correct sizes, we compile the set of furniture items as $F = \{f_1, \dots, f_n\}$.

\subsection{Object Placement}

The generation of a layout can be defined as follows: given a set of furniture items $F = \{f_1, \dots, f_n\}$, where each item $f_i$ is characterized by its identifier $Objkey(f_i)$ and bounding box ${Rect}(f_i)$, the goal is to determine a placement 
$P = \{(x_1, y_1, z_1, \theta_1), \dots, (x_n, y_n, z_n, \theta_n)\}$ for all items. Here, $(x_i, y_i, z_i)$ and $\theta_i$ denote the position and orientation of the item $f_i$, respectively.
Unlike previous works that primarily focus on placing objects on the floor, 
our approach allows for arbitrary placement on or under other objects, or on the wall. We categorize $F$ into three placement types, $F_{floor}$, $F_{wall}$, and $F_{other}$, through interaction with the MLLM agent, following the vision-question paradigm.
Instead of directly predicting the layout plan in a single interaction like \textit{LayoutGPT} \cite{feng2024layoutgpt} and \textit{AnyHome} \cite{wen2023anyhome}, we employ Chain-of-Thought (CoT) prompting to systematically reasons through the plan step by step, which leads to a more precise process.

\noindent{\textbf{Floor Placement.}} 
We first arrange items in $F_{floor}$ using the Grid Coverage Algorithm in Figure \ref{fig:method_grd}. We overlay a grid on the floor, where each cell is identified by a unique alphanumeric ID. Assuming that walls are along the top and left edges of the floor, we label wall-adjacent cells as `wall'. We then ask the agent to specify the grid cells that each furniture item in $F_{floor}$ should cover.
Since directly assigning cells for a large set of items can be challenging, we propose a local by local placement method (Figure \ref{fig:method_grd}-(a)), which stages the larger items before smaller ones, leading to simplified tasks and improved performance.
For very large objects that can cause inaccurate coverage predictions, we propose a scale-adaptive grid placement method, which initially place them on a sparse grid to minimize the number of covered cells (Figure \ref{fig:method_grd}-(b)-left) and then project them onto a denser grid for further placement of additional items (Figure \ref{fig:method_grd}-(b)-right).

\noindent{\textbf{3D Placement.}}
After placing $F_{floor}$, we apply the same grid coverage strategy to $F_{wall}$ by similarly overlaying a grid on the wall. In addition, for all items in $F_{floor}$ that are adjacent to the wall, we project them to the wall grid before placing $F_{wall}$ to avoid collision between these two types of items.
We then apply the same strategy to place $F_{other}$, such as putting a laptop on top of a table. In this way, our layout generation method is highly versatile and adaptable to irregular floor plans, with or without walls.

\begin{figure}[t]
\centering
\includegraphics[width=0.98\linewidth]{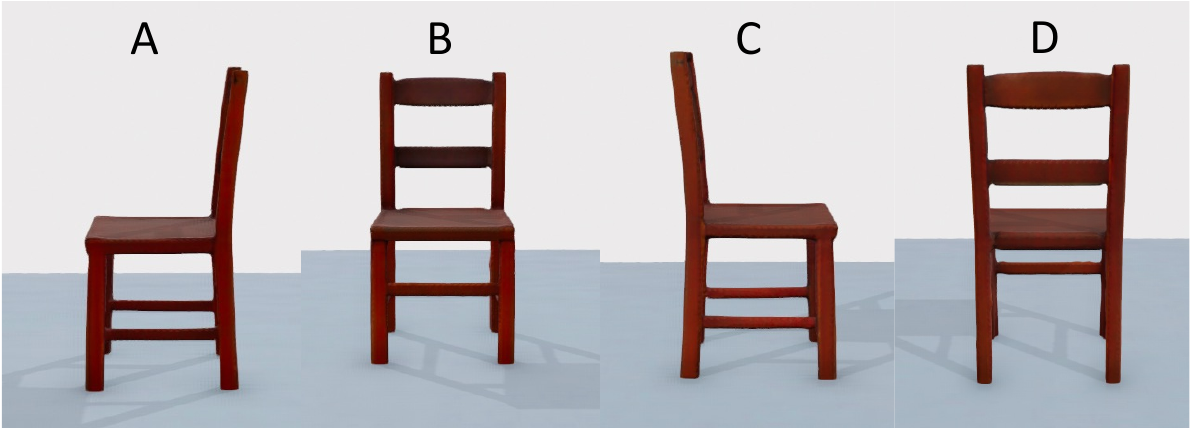}
\caption{\textbf{A visual content for initial orientation prediction.} The agent will select the initial orientation of the chair from four views.}
\label{fig:method_orien}
\end{figure}

\noindent{\textbf{Orientation Prediction.}}
After placing all furniture items, we further correct their orientations. The aforementioned pose alignment technique ensures that the frontal face of each item aligns with one of four views: left, right, front, or back. The agent only needs to identify the frontal face from the four candidate views (Figure \ref{fig:method_orien}).
After that, we mark the orientation of the frontal face on the grid with a red arrow (Figure \ref{fig:method_grd}). Finally, another chat session is then used to determine the rotation angle for the final face orientation.

\subsection{Self-Reflection}

Self-reflection empowers agents to iteratively improve by refining decisions and correcting past errors \cite{yao2023react,shinn2024reflexion}. To this end, we incorporate a visual self-reflection mechanism into our agent system. After all furniture items are placed, we start a continuous chat by asking whether the result meets user requirements, and is reasonable and realistic, allowing the agent to visually inspect the arrangement within the 3D interface. If the placement does not satisfy user requirements or violates realistic and rational constraints, the agent is prompted to adjust the layout accordingly. Furthermore, we integrate surface collision detection during object placement. This is particularly important for irregularly shaped objects, as it helps prevent collisions or floating objects.

\section{Offline-to-Online Reference Search}
\label{subsec:o2o}

To provide references for the visual-question paradigm, we propose an Offline-to-Online Search method (O2O-Search) to identify the minimal informative support set.
 
Given a database containing $N$ examples $D = \{ (X_i, Y_i) \}_{i=1}^N$, where each input set $X_i = (u_i, t_i, v_i)$ consists of a user text prompt $u_i$, an indoor scene text description $t_i$, and visual captures of the scene $v_i$, and $Y_i$ represents the corresponding output result, all in JSON format. Our O2O-Search method aims to automatically identify the minimal informative support set within this dataset.

\noindent{\textbf{Offline Grouping.}} 
We observe that each example $D_i=(X_i, Y_i)$ can be categorized based on its characteristics. To achieve this, we assign a category identifier to each example in $D$, effectively dividing it into distinct groups. Within each group, we measure the similarity between any two examples $(D_i, D_j)$ using the following criteria:
\begin{equation}
    f(D_i,D_j)=D_\text{CLIP}(u_i,u_j)+\alpha D_\text{CLIP}(t_i,t_j)+\beta D_\text{DINO}(v_i,v_j),
\label{eq:sim}
\end{equation}
where $D_\text{CLIP}$ is the CLIP cosine similarity \cite{radford2021learning} and $D_\text{DINO}$ is the DINOv2 cosine similarity \cite{oquab2023dinov2}.
The information entropy of an example $D_i$ is defined as:
\begin{equation}
    E(D_i)={\textstyle\sum}_{j=0,j\neq i}^{N}f(D_i,D_j).
\end{equation}
We calculate the information entropy for each group and select the top-$K$ demonstrations with the highest information entropy values within each group.

\noindent{\textbf{Online Retrieval.}} 
Given a test input $X_t$ during user interaction, we ask the agent to assign a category identifier. We then select the group corresponding to this identifier as the final supporting set, providing the most relevant reference examples for the current task.

\section{Experiments}

In this section, we detail the evaluation setup, present ablation studies and comparisons to baselines, and showcase diverse results generated by out method. For further implementation details and additional evaluations, please refer to our supplementary materials.

\subsection{Evaluation Setup}
\label{subsec:es}

\begin{figure*}
\centering 
\includegraphics[width=0.87\textwidth]{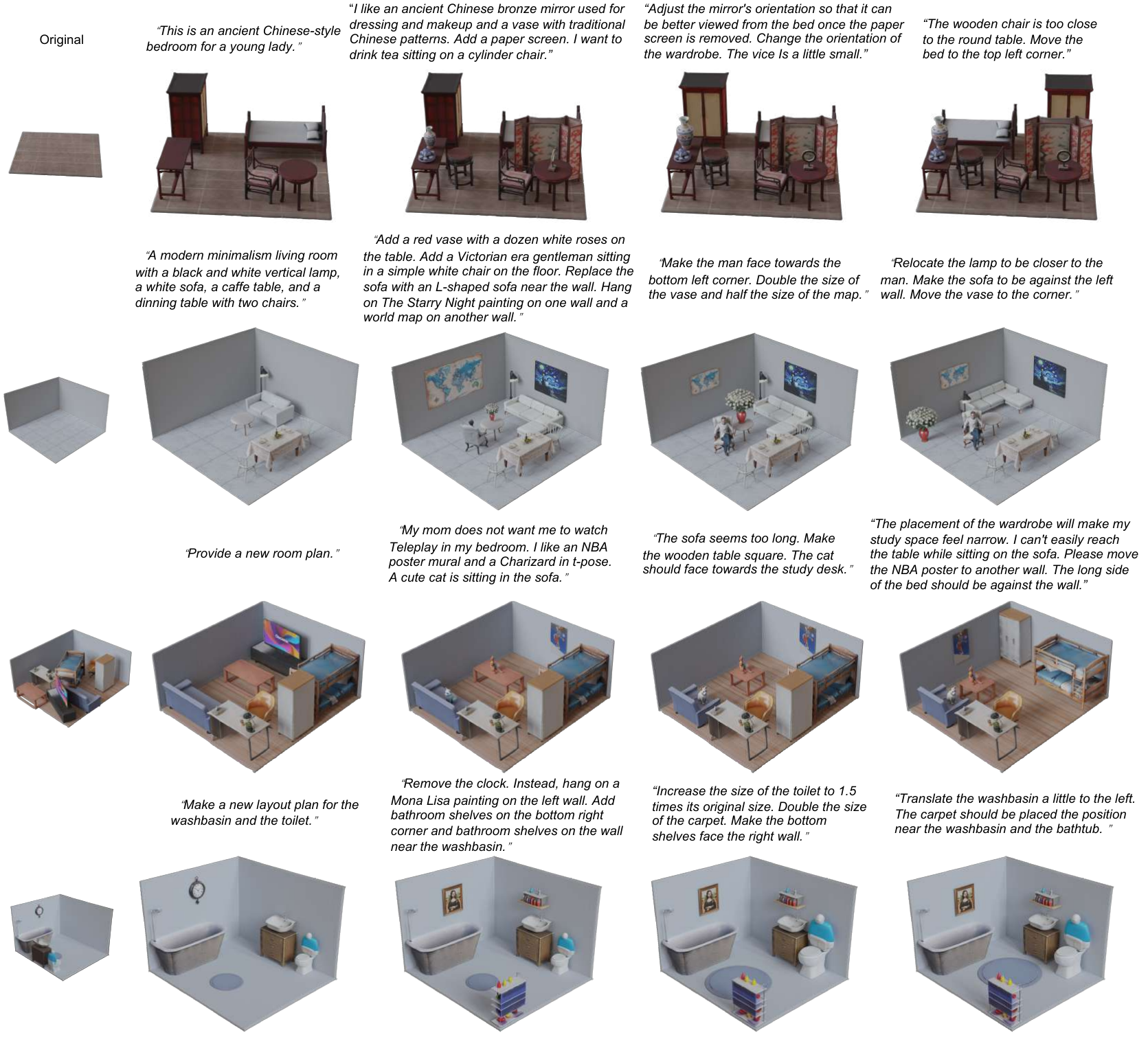}
\caption{\textbf{Interactive Layout Generation Results.} \name facilitates multi-turn conversational user interactions for layout design, a feature unsupported by other methods.}
\label{fig:result}
\end{figure*}

To evaluate the quality of our layout generation, we first collected a set of various real scenes online and wrote corresponding text descriptions for each. We then generated the dataset by producing 10 layout plans for each scene, similar to cases 1, 2, and 7 in Figure \ref{fig:result}. We then conduct a comparative analysis using the following metrics:

\noindent\textbf{Out of Bound Rate (OOB).} This measures the percentage of layout plans where objects extend beyond the room's boundaries or intersect with other objects, indicating the quality of spatial arrangement.
We report OOB at the layout plan level, 
checking for the presence of these issues with each plan, following \textit{LayoutGPT} and \textit{AnyHome}.

\noindent\textbf{Orientation Correctness (ORI).} This evaluates the correctness of object orientations within the layout context.
Correct orientation is defined as the object being easily accessible and usable by users in the room.
For example, a wardrobe adjacent to but facing a wall is considered incorrectly oriented.
We manually check each layout plan and report the percentage of layout plans where furniture has no unreasonable face orientations.

\noindent\textbf{CLIP Similarity (CLIP-Sim).} This assesses the alignment between the user's text description and the scene content by calculating the text-to-image CLIP similarity, using multi-view renderings of the 3D layout scene.
Random viewpoint perturbations are incorporated to enhance the robustness of our evaluations.

\subsection{Ablation Studies}

\begin{table}[]
\centering
\begin{tabular}{c|ccc}
\hline
 & OOB $\downarrow$ & ORI $\uparrow$ & CLIP-Sim $\uparrow$ \\ \hline
InstructScene & 38.8 & 73.3 & 13.1 \\
LayoutGPT & 52.8 & 61.3 & 18.4 \\
AnyHome & 32.8 & 72.3 & 23.2 \\ \hline
Ours-Visual & 36.8 & 63.8 & 20.1 \\
Ours-w/o VTP & 37.5 & 62.0 & 19.6 \\
Ours-w/o O2O-Search & 31.8 & 75.3 & 21.9 \\
Ours-w/o SR & 24.8 & 81.0 & 24.7 \\ \hline
Ours & \textbf{21.0} & \textbf{84.8} & \textbf{27.1} \\ \hline
\end{tabular}
\caption{
\textbf{Layout quality comparison.} Our method achieves the best performance with a smaller Out-of-Boundary Rate (OOB), better Orientation Correction (ORI), and higher CLIP Similarity (CLIP-Sim).
}
\label{tab:abl_cmp}
\end{table}

\begin{figure}[t]
\centering
\includegraphics[width=0.90\linewidth]{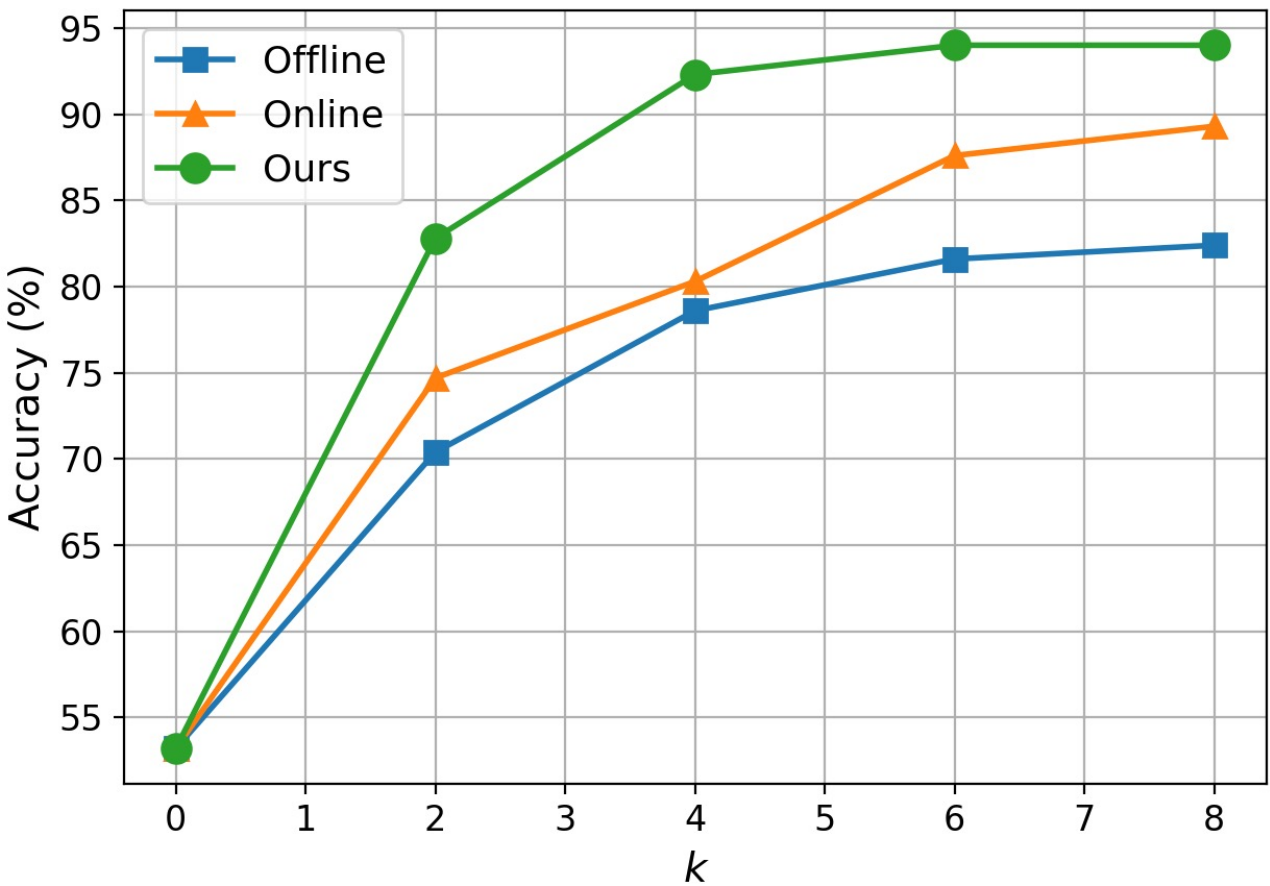}
\caption{\textbf{Ablation study for reference search methods.} Our O2O-Search method achieves the best performance compared to other baselines on different reference size $k$.}
\label{fig:abl_ICL}
\end{figure}

\begin{figure*}
\centering
\includegraphics[width=0.86\linewidth]{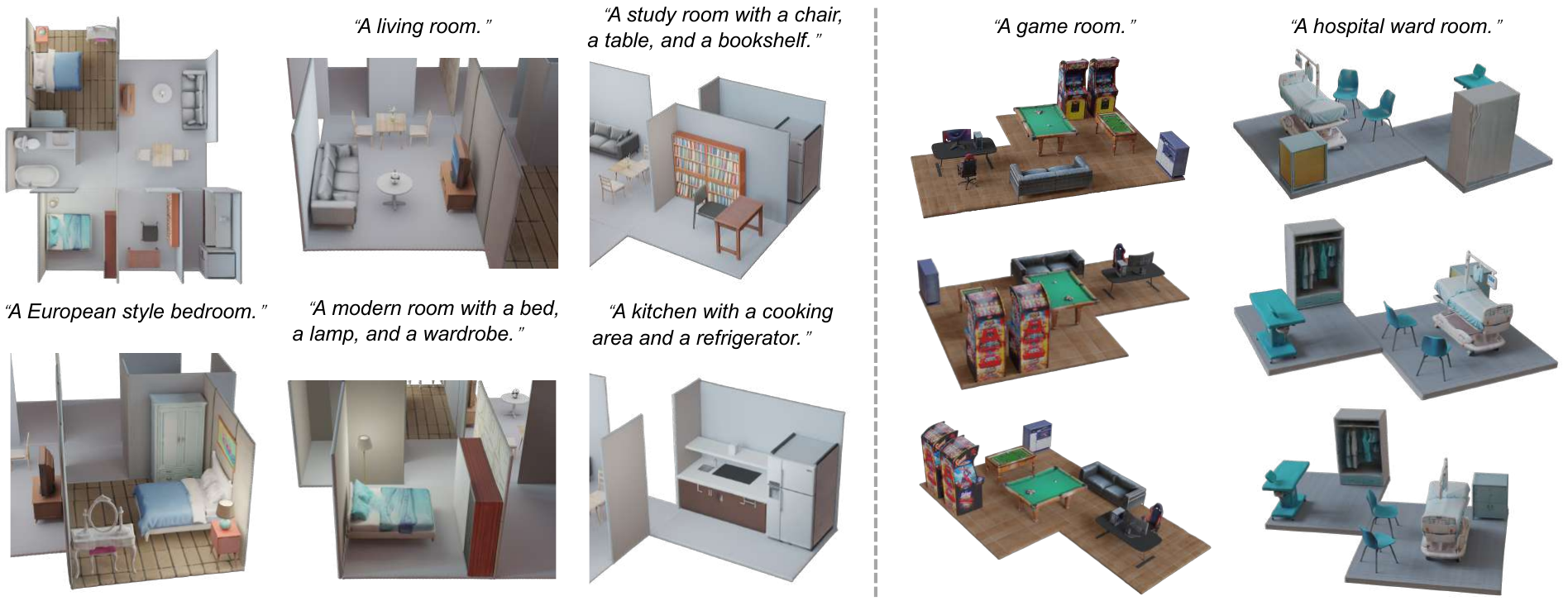}
\caption{\textbf{Extended layout generation results.} \name can handle multi-room setups and irregular floor plans.}
\label{fig:ext}
\end{figure*}

\begin{figure*}
\centering 
\includegraphics[width=0.81\textwidth]{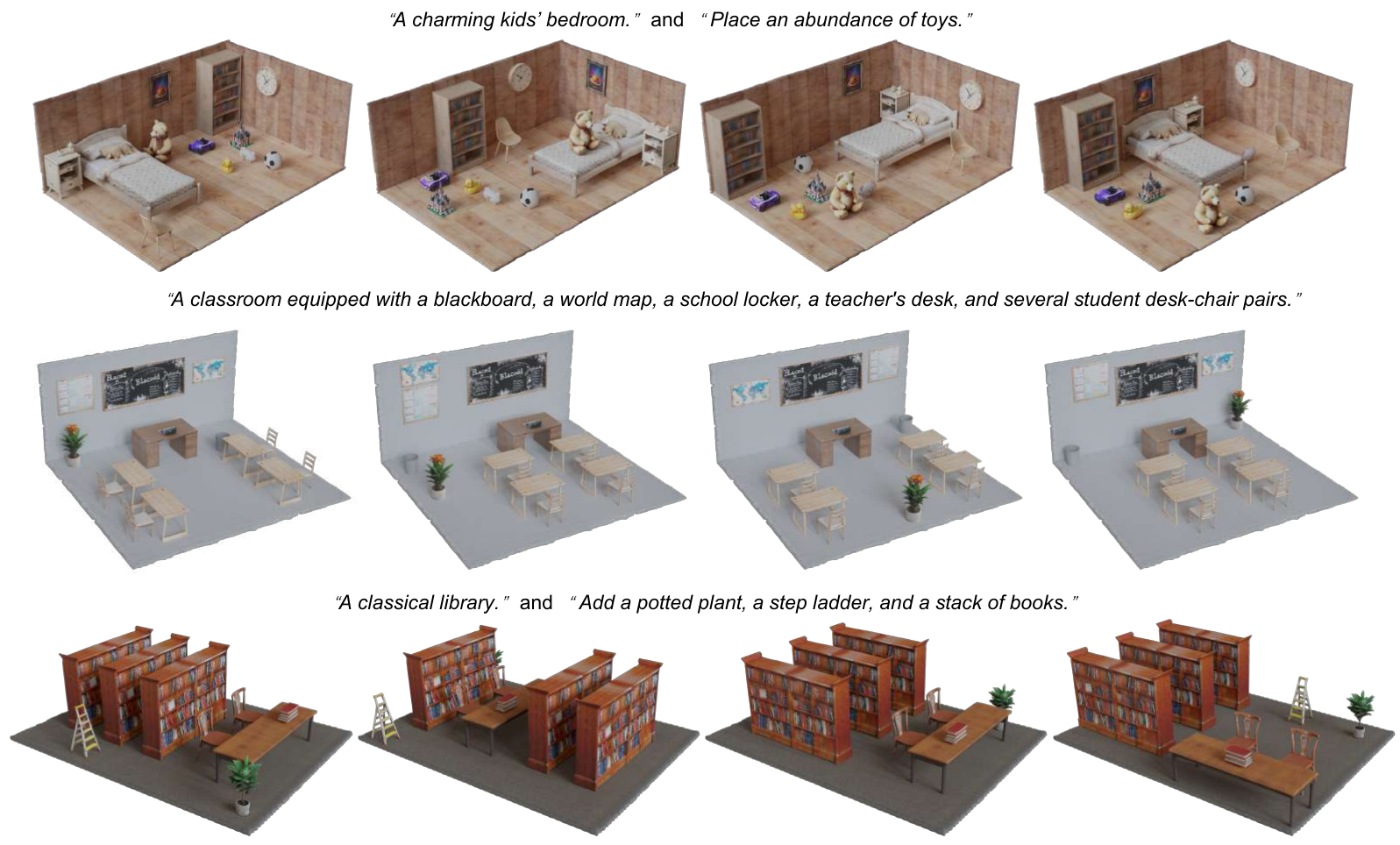}
\caption{\textbf{Diverse layout generation results.} \name enables the generation of diverse layout plans for a same scene.}
\label{fig:diverse}
\end{figure*}

\begin{figure*}
\centering 
\includegraphics[width=0.81\textwidth]{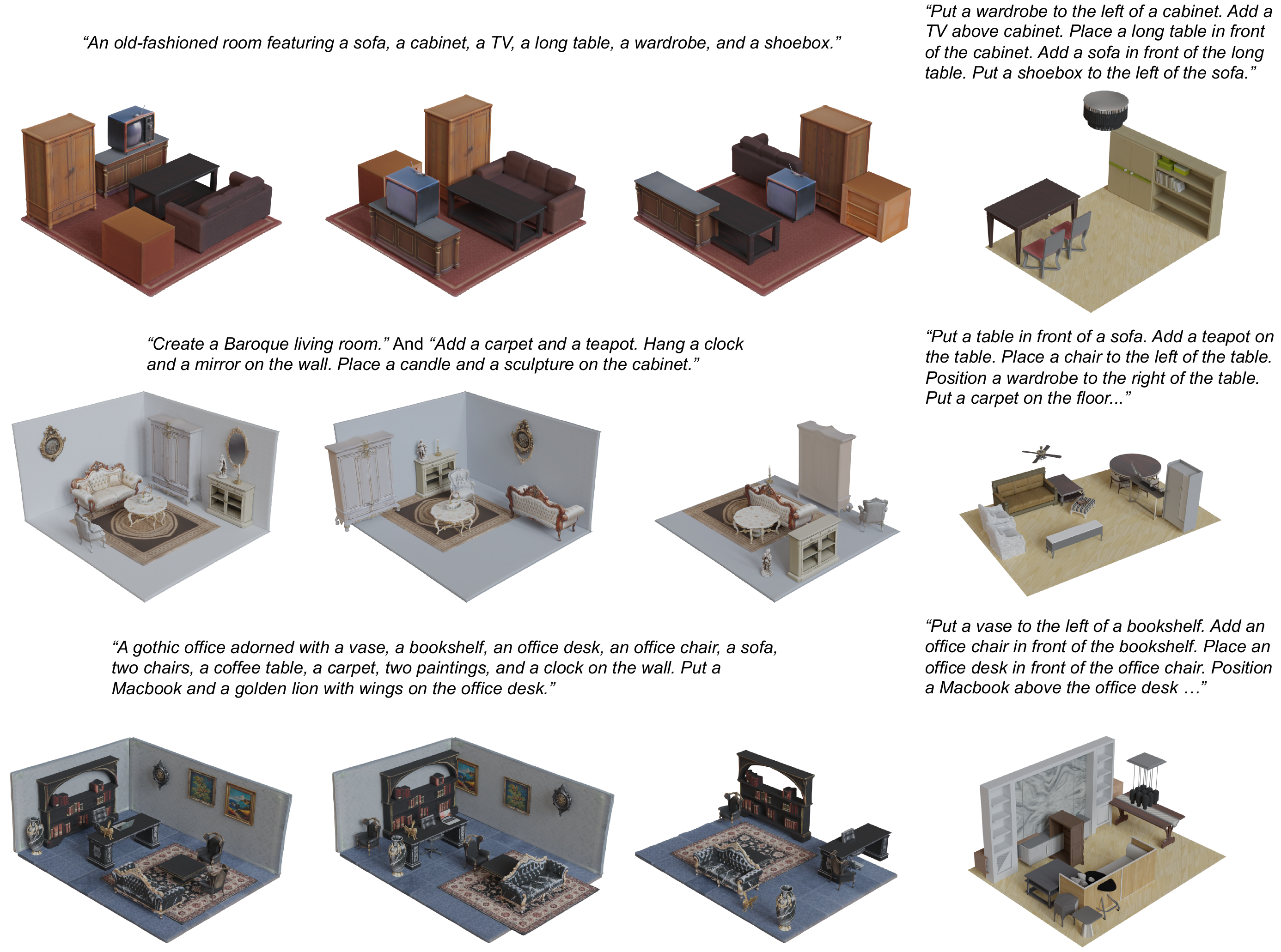}
\begin{small}
\begin{tabular}{p{0.10\linewidth}
<{\centering}p{0.17\linewidth}
<{\centering}p{0.17\linewidth}
<{\centering}p{0.17\linewidth}
<{\centering}p{0.17\linewidth}
<{\centering}p{0.24\linewidth}<{\centering} }
&Ours & AnyHome & LayoutGPT & InstructScene &
\end{tabular}
\end{small}
\caption{\textbf{Comparison.} Our method produces more reasonable layout plans compared to other methods.}
\label{fig:cmp}
\end{figure*}

\noindent\textbf{Visual-Text Prompting.}
In Table \ref{tab:abl_cmp}, we evaluate the effectiveness of visual-text prompting. Results demonstrate that visual-text prompting significantly enhances layout generation compared to the baseline of \textit{Ours-w/o-VTP}, which relies solely on text descriptions. This is evidenced by a lower OBB, better ORI, and higher CLIP-Sim scores.

We also observe a marginal improvement in the baseline of \textit{Ours-Visual}, which directly integrates visual captures of the environment without visual-text prompting, compared to \textit{Ours-w/o-VTP}. This suggests that while visual information is beneficial, its inclusion alone without structural integration does not lead to substantial performance gains. In contrast, our visual-text prompting technique enables more effective utilization of visual information, contributing to better overall performance.

Furthermore, our full method incorporating the self-reflection mechanism further enhances layout generation performance compared to the baseline of \textit{Ours-w/o-SR} by providing feedback to verify and refine results.

\noindent\textbf{O2O-Search.}
We also conduct an ablation study to assess the effectiveness of our O2O-Search method. We randomly select $30\%$ of the questions from our example dataset $D$ to form the test set. Our method is compared against two baselines: 1) An offline-only method that calculates pairwise similarity across the entire dataset and selects the top-$k$ exemplars without grouping \cite{su2022selective}; and 2) An online-only method that computes similarity between the test prompt and all candidates in the dataset to select supporting references \cite{alayrac2022flamingo,yang2022empirical,feng2024layoutgpt}.

We present the accuracy across various sizes of the support set $k$ in Figure \ref{fig:abl_ICL}. When $k=0$, meaning no support set, the accuracy remains significantly low, underscoring the important role that support references play. As we increase the value of $k$, there is a general trend of improved accuracy for all baselines;. However, this improvement begin to consistently plateau around $k = 8$, suggesting diminishing returns on accuracy gains for larger $k$. This indicates that there is an optimal range of $k$ that maximizes accuracy without incurring too much computational cost due to excessive size of references. Thus, we use $k=8$ in our method.
Importantly, our O2O-Search method consistently outperforms both the offline and online baselines in accuracy across all values of $k$, demonstrating its superior scalability and effectiveness in utilizing larger support sets to boost performance.

We also observe a substantial degradation in OBB, ORI, and CLIP-Sim metrics when O2O-Search is removed, as shown in Table \ref{tab:abl_cmp}. This further highlights the curcial role that our O2O-Search plays in enhancing the overall performance of our system.

\subsection{Comparisons}

We compare \name with three existing methods: \textit{InstructScene} \cite{lin2023instructscene}, \textit{LayoutGPT} \cite{feng2024layoutgpt}, and \textit{AnyHome} \cite{feng2024layoutgpt}, with \textit{AnyHome} being a concurrent work. 
For a fair comparison, we use the same furniture set generated by our model, with identical scale values and initial orientations, for \textit{layoutGPT} and \textit{AnyHome}. 
Moreover, for \textit{InstructScene}, we provide a detailed description of our generated layout plan to obtain their results.
It should be noted that in our method, 
the furniture is generated using a text-to-3D API, while \textit{layoutGPT} and \textit{AnyHome} require searching a large dataset. But here We use the same set generated by ours for a fair layout quality comparison.

\noindent\textbf{Quantitative Comparison.}
Table \ref{tab:abl_cmp} provides the quantitative comparison. 
\textit{InstructScene} demonstrates weaker performance in terms of ORI and CLIP-Sim compared to all LLM-based methods. 
\textit{LayoutGPT}, while solely leveraging an online-only support set without visual input, struggles to produce satisfactory results.
\textit{AnyHome} attempts to address the issues of \textit{LayoutGPT} by incorporating a set of manually defined placement rules in texts.
However, it only uses visual information for furniture position and pose refinement through SDS optimization, limiting its overall effectiveness.
In contrast, \name achieve the best results on all metrics, thanks to the integration of visual input and a more advanced support set search method.

\noindent\textbf{Qualitative Comparison.}
Figure \ref{fig:cmp} presents the qualitative comparison.
Compared to our method, other approaches are more susceptible to issues like out-of-bound placements, incorrect orientations, and poor content alignments.
Specifically, \textit{InstructScene} is limited to concrete text instructions detailing spatial relations and may generate non-existent objects due to text-to-graph conversion errors. 
Both \textit{InstructScene} and \textit{LayoutGPT} lack support for wall placements such as paintings.
\textit{AnyHome}, while outperforming the other two, relies on pre-defined human-knowledge rules, leading to illogical placements, such as neglecting common-sense spacing between objects.
In addition, all three methods are prone to boundary and orientation issues, impacting usability and aesthetics.
In contrast, \name demonstrates superior performance by fully exploiting visual information. 
Moreover, \name supports multi-turn conventional interactions, as demonstrated in Figure \ref{fig:result}, a feature absent in other methods.

\subsection{More Visual Results}

We present our interactive layout generation results in Figure \ref{fig:result}, a feature unsupported by other methods. \name empowers users to create layouts from scratch, transforming an empty space into diverse designs such as an ancient Chinese-style bedroom or a modern minimalist living room. Users can also perform partial rearrangements, as illustrated in the washing room scenario. \name supports precise 3D placements, such as positioning a vase on a table in case one or hanging a painting on a wall in case two. Additionally, users can modify layouts through instructions, including inserting new items, removing objects, and adjusting placements in terms of scale, rotation, and translation. Notably, \name excels at handling open-vocabulary instructions, accommodating both abstract and concrete descriptions, whether mixed or focused.

\name supports diverse layout generation by prompting the agent to provide multiple layouts for the same set of furniture items. As illustrated in Figure \ref{fig:diverse}, we show three cases, each with four distinct layout plans. \name enables the creation of varied layout plans, accommodating objects on both floors and walls, while ensuring that the orientations of objects are not only diverse but also logical within the context of the room and its functionality.

Figure \ref{fig:ext} illustrates the versatility and adaptability of our method in handling a variety of layout configurations, including irregular floor shapes and multi-room setups. For instance, we employ an irregular grid to generate a layout for an L-shaped game room. In multi-room scenarios, we sequentially generate each individual room, demonstrating the system's capability in managing complex spatial arrangements across multiple interconnected spaces.

\section{Conclusion}

In this paper, we introduce \textit{Chat2Layout}, a novel method to indoor layout generation that expands the capabilities of MLLMs beyond vision interpretation and text generation. Our method establishes a unified vision-question paradigm for in-context learning, enabling communication with the MLLM to guide its reasoning with both textual and visual information. Within this framework, we introduce a training-free visual prompting technique that facilitates the creation of realistic and contextually appropriate layouts. In addition, we employ an O2O-Search method to identify the minimal support set as references for effective prompting. Overall, our method interprets and handles user requirements and 3D environments both linguistically and visually and execute challenging furniture arrangement tasks. This enables diverse and complex 3D layout generation, providing interactive user experiences that surpass previous methods.

\bibliography{reference}

\begin{thebibliography}{80}
\providecommand{\natexlab}[1]{#1}

\bibitem[{Alayrac et~al.(2022)Alayrac, Donahue, Luc, Miech, Barr, Hasson, Lenc, Mensch, Millican, Reynolds et~al.}]{alayrac2022flamingo}
Alayrac, J.-B.; Donahue, J.; Luc, P.; Miech, A.; Barr, I.; Hasson, Y.; Lenc, K.; Mensch, A.; Millican, K.; Reynolds, M.; et~al. 2022.
\newblock Flamingo: a visual language model for few-shot learning.
\newblock \emph{Advances in neural information processing systems}, 35: 23716--23736.

\bibitem[{Belzner, Gabor, and Wirsing(2023)}]{belzner2023large}
Belzner, L.; Gabor, T.; and Wirsing, M. 2023.
\newblock Large language model assisted software engineering: prospects, challenges, and a case study.
\newblock In \emph{International Conference on Bridging the Gap between AI and Reality}, 355--374. Springer.

\bibitem[{Chan et~al.(2023)Chan, Chen, Su, Yu, Xue, Zhang, Fu, and Liu}]{chan2023chateval}
Chan, C.-M.; Chen, W.; Su, Y.; Yu, J.; Xue, W.; Zhang, S.; Fu, J.; and Liu, Z. 2023.
\newblock Chateval: Towards better llm-based evaluators through multi-agent debate.
\newblock \emph{arXiv preprint arXiv:2308.07201}.

\bibitem[{Chang et~al.(2015)Chang, Monroe, Savva, Potts, and Manning}]{chang2015text}
Chang, A.; Monroe, W.; Savva, M.; Potts, C.; and Manning, C.~D. 2015.
\newblock Text to 3D Scene Generation with Rich Lexical Grounding.
\newblock In \emph{Proceedings of the 53rd Annual Meeting of the Association for Computational Linguistics and the 7th International Joint Conference on Natural Language Processing (Volume 1: Long Papers)}, 53--62.

\bibitem[{Chang, Savva, and Manning(2014)}]{chang2014learning}
Chang, A.; Savva, M.; and Manning, C.~D. 2014.
\newblock Learning spatial knowledge for text to 3D scene generation.
\newblock In \emph{Proceedings of the 2014 conference on empirical methods in natural language processing (EMNLP)}, 2028--2038.

\bibitem[{Chen et~al.(2024)Chen, Wang, Tian, Ye, Gao, Cui, Tong, Hu, Luo, Ma et~al.}]{chen2024far}
Chen, Z.; Wang, W.; Tian, H.; Ye, S.; Gao, Z.; Cui, E.; Tong, W.; Hu, K.; Luo, J.; Ma, Z.; et~al. 2024.
\newblock How Far Are We to GPT-4V? Closing the Gap to Commercial Multimodal Models with Open-Source Suites.
\newblock \emph{arXiv preprint arXiv:2404.16821}.

\bibitem[{Chu et~al.(2024)Chu, Qiao, Zhang, Xu, Wei, Yang, Sun, Hu, Lin, Zhang et~al.}]{chu2024mobilevlm}
Chu, X.; Qiao, L.; Zhang, X.; Xu, S.; Wei, F.; Yang, Y.; Sun, X.; Hu, Y.; Lin, X.; Zhang, B.; et~al. 2024.
\newblock MobileVLM V2: Faster and Stronger Baseline for Vision Language Model.
\newblock \emph{arXiv preprint arXiv:2402.03766}.

\bibitem[{Deitke et~al.(2023)Deitke, Schwenk, Salvador, Weihs, Michel, VanderBilt, Schmidt, Ehsani, Kembhavi, and Farhadi}]{deitke2023objaverse}
Deitke, M.; Schwenk, D.; Salvador, J.; Weihs, L.; Michel, O.; VanderBilt, E.; Schmidt, L.; Ehsani, K.; Kembhavi, A.; and Farhadi, A. 2023.
\newblock Objaverse: A universe of annotated 3d objects.
\newblock In \emph{Proceedings of the IEEE/CVF Conference on Computer Vision and Pattern Recognition}, 13142--13153.

\bibitem[{Deng et~al.(2024)Deng, Gu, Zheng, Chen, Stevens, Wang, Sun, and Su}]{deng2024mind2web}
Deng, X.; Gu, Y.; Zheng, B.; Chen, S.; Stevens, S.; Wang, B.; Sun, H.; and Su, Y. 2024.
\newblock Mind2web: Towards a generalist agent for the web.
\newblock \emph{Advances in Neural Information Processing Systems}, 36.

\bibitem[{Dhamo et~al.(2021)Dhamo, Manhardt, Navab, and Tombari}]{dhamo2021graph}
Dhamo, H.; Manhardt, F.; Navab, N.; and Tombari, F. 2021.
\newblock Graph-to-3d: End-to-end generation and manipulation of 3d scenes using scene graphs.
\newblock In \emph{Proceedings of the IEEE/CVF International Conference on Computer Vision}, 16352--16361.

\bibitem[{Dong et~al.(2024)Dong, Zhang, Zang, Cao, Wang, Ouyang, Zhang, Duan, Zhang, Li et~al.}]{dong2024internlm}
Dong, X.; Zhang, P.; Zang, Y.; Cao, Y.; Wang, B.; Ouyang, L.; Zhang, S.; Duan, H.; Zhang, W.; Li, Y.; et~al. 2024.
\newblock InternLM-XComposer2-4KHD: A Pioneering Large Vision-Language Model Handling Resolutions from 336 Pixels to 4K HD.
\newblock \emph{arXiv preprint arXiv:2404.06512}.

\bibitem[{Fabri and Pion(2009)}]{fabri2009cgal}
Fabri, A.; and Pion, S. 2009.
\newblock CGAL: The computational geometry algorithms library.
\newblock In \emph{Proceedings of the 17th ACM SIGSPATIAL international conference on advances in geographic information systems}, 538--539.

\bibitem[{Feng et~al.(2024)Feng, Zhu, Fu, Jampani, Akula, He, Basu, Wang, and Wang}]{feng2024layoutgpt}
Feng, W.; Zhu, W.; Fu, T.-j.; Jampani, V.; Akula, A.; He, X.; Basu, S.; Wang, X.~E.; and Wang, W.~Y. 2024.
\newblock Layoutgpt: Compositional visual planning and generation with large language models.
\newblock \emph{Advances in Neural Information Processing Systems}, 36.

\bibitem[{Fisher et~al.(2012)Fisher, Ritchie, Savva, Funkhouser, and Hanrahan}]{fisher2012example}
Fisher, M.; Ritchie, D.; Savva, M.; Funkhouser, T.; and Hanrahan, P. 2012.
\newblock Example-based synthesis of 3D object arrangements.
\newblock \emph{ACM Transactions on Graphics (TOG)}, 31(6): 1--11.

\bibitem[{Fu et~al.(2021{\natexlab{a}})Fu, Cai, Gao, Zhang, Wang, Li, Zeng, Sun, Jia, Zhao et~al.}]{fu20213d}
Fu, H.; Cai, B.; Gao, L.; Zhang, L.-X.; Wang, J.; Li, C.; Zeng, Q.; Sun, C.; Jia, R.; Zhao, B.; et~al. 2021{\natexlab{a}}.
\newblock 3d-front: 3d furnished rooms with layouts and semantics.
\newblock In \emph{Proceedings of the IEEE/CVF International Conference on Computer Vision}, 10933--10942.

\bibitem[{Fu et~al.(2021{\natexlab{b}})Fu, Jia, Gao, Gong, Zhao, Maybank, and Tao}]{fu20213df}
Fu, H.; Jia, R.; Gao, L.; Gong, M.; Zhao, B.; Maybank, S.; and Tao, D. 2021{\natexlab{b}}.
\newblock 3d-future: 3d furniture shape with texture.
\newblock \emph{International Journal of Computer Vision}, 129: 3313--3337.

\bibitem[{Fu et~al.(2017)Fu, Chen, Wang, Wen, Zhou, and Fu}]{fu2017adaptive}
Fu, Q.; Chen, X.; Wang, X.; Wen, S.; Zhou, B.; and Fu, H. 2017.
\newblock Adaptive synthesis of indoor scenes via activity-associated object relation graphs.
\newblock \emph{ACM Transactions on Graphics (TOG)}, 36(6): 1--13.

\bibitem[{Gravitas(2023)}]{AutoGPT}
Gravitas, S. 2023.
\newblock AutoGPT.
\newblock \url{https://agpt.co}.

\bibitem[{Haque et~al.(2023)Haque, Tancik, Efros, Holynski, and Kanazawa}]{haque2023instruct}
Haque, A.; Tancik, M.; Efros, A.~A.; Holynski, A.; and Kanazawa, A. 2023.
\newblock Instruct-nerf2nerf: Editing 3d scenes with instructions.
\newblock In \emph{Proceedings of the IEEE/CVF International Conference on Computer Vision}, 19740--19750.

\bibitem[{Hart(2006)}]{hart2006nasa}
Hart, S.~G. 2006.
\newblock NASA-task load index (NASA-TLX); 20 years later.
\newblock In \emph{Proceedings of the human factors and ergonomics society annual meeting}, volume~50, 904--908. Sage publications Sage CA: Los Angeles, CA.

\bibitem[{Hong et~al.(2024)Hong, Tang, Cao, Shi, Wu, Chen, Wang, Pan, Lin, and Liu}]{hong20243dtopia}
Hong, F.; Tang, J.; Cao, Z.; Shi, M.; Wu, T.; Chen, Z.; Wang, T.; Pan, L.; Lin, D.; and Liu, Z. 2024.
\newblock 3DTopia: Large Text-to-3D Generation Model with Hybrid Diffusion Priors.
\newblock \emph{arXiv preprint arXiv:2403.02234}.

\bibitem[{Lester, Al-Rfou, and Constant(2021)}]{lester2021power}
Lester, B.; Al-Rfou, R.; and Constant, N. 2021.
\newblock The Power of Scale for Parameter-Efficient Prompt Tuning.
\newblock In \emph{Proceedings of the 2021 Conference on Empirical Methods in Natural Language Processing}, 3045--3059.

\bibitem[{Li et~al.(2024{\natexlab{a}})Li, Wong, Zhang, Usuyama, Liu, Yang, Naumann, Poon, and Gao}]{li2024llava}
Li, C.; Wong, C.; Zhang, S.; Usuyama, N.; Liu, H.; Yang, J.; Naumann, T.; Poon, H.; and Gao, J. 2024{\natexlab{a}}.
\newblock Llava-med: Training a large language-and-vision assistant for biomedicine in one day.
\newblock \emph{Advances in Neural Information Processing Systems}, 36.

\bibitem[{Li et~al.(2023)Li, Tan, Zhang, Xu, Luan, Xu, Hong, Sunkavalli, Shakhnarovich, and Bi}]{li2023instant3d}
Li, J.; Tan, H.; Zhang, K.; Xu, Z.; Luan, F.; Xu, Y.; Hong, Y.; Sunkavalli, K.; Shakhnarovich, G.; and Bi, S. 2023.
\newblock Instant3D: Fast Text-to-3D with Sparse-view Generation and Large Reconstruction Model.
\newblock In \emph{The Twelfth International Conference on Learning Representations}.

\bibitem[{Li et~al.(2019)Li, Patil, Xu, Chaudhuri, Khan, Shamir, Tu, Chen, Cohen-Or, and Zhang}]{li2019grains}
Li, M.; Patil, A.~G.; Xu, K.; Chaudhuri, S.; Khan, O.; Shamir, A.; Tu, C.; Chen, B.; Cohen-Or, D.; and Zhang, H. 2019.
\newblock Grains: Generative recursive autoencoders for indoor scenes.
\newblock \emph{ACM Transactions on Graphics (TOG)}, 38(2): 1--16.

\bibitem[{Li et~al.(2024{\natexlab{b}})Li, Zhang, Wang, Zhong, Chen, Chu, Liu, and Jia}]{li2024mini}
Li, Y.; Zhang, Y.; Wang, C.; Zhong, Z.; Chen, Y.; Chu, R.; Liu, S.; and Jia, J. 2024{\natexlab{b}}.
\newblock Mini-Gemini: Mining the Potential of Multi-modality Vision Language Models.
\newblock \emph{arXiv preprint arXiv:2403.18814}.

\bibitem[{Lin and Yadong(2023)}]{lin2023instructscene}
Lin, C.; and Yadong, M. 2023.
\newblock InstructScene: Instruction-Driven 3D Indoor Scene Synthesis with Semantic Graph Prior.
\newblock In \emph{The Twelfth International Conference on Learning Representations}.

\bibitem[{Lin et~al.(2023)Lin, Gao, Tang, Takikawa, Zeng, Huang, Kreis, Fidler, Liu, and Lin}]{lin2023magic3d}
Lin, C.-H.; Gao, J.; Tang, L.; Takikawa, T.; Zeng, X.; Huang, X.; Kreis, K.; Fidler, S.; Liu, M.-Y.; and Lin, T.-Y. 2023.
\newblock Magic3d: High-resolution text-to-3d content creation.
\newblock In \emph{Proceedings of the IEEE/CVF Conference on Computer Vision and Pattern Recognition}, 300--309.

\bibitem[{LumaAI(2023)}]{Luma}
LumaAI. 2023.
\newblock Luma AI.
\newblock \url{https://lumalabs.ai/}.

\bibitem[{Luo et~al.(2020)Luo, Zhang, Wu, and Tenenbaum}]{luo2020end}
Luo, A.; Zhang, Z.; Wu, J.; and Tenenbaum, J.~B. 2020.
\newblock End-to-end optimization of scene layout.
\newblock In \emph{Proceedings of the IEEE/CVF Conference on Computer Vision and Pattern Recognition}, 3754--3763.

\bibitem[{Ma et~al.(2016)Ma, Li, Zou, Liao, Tong, and Zhang}]{ma2016action}
Ma, R.; Li, H.; Zou, C.; Liao, Z.; Tong, X.; and Zhang, H. 2016.
\newblock Action-driven 3D indoor scene evolution.
\newblock \emph{ACM Transactions on Graphics (TOG)}, 35(6): 1--13.

\bibitem[{Ma et~al.(2018)Ma, Patil, Fisher, Li, Pirk, Hua, Yeung, Tong, Guibas, and Zhang}]{ma2018language}
Ma, R.; Patil, A.~G.; Fisher, M.; Li, M.; Pirk, S.; Hua, B.-S.; Yeung, S.-K.; Tong, X.; Guibas, L.; and Zhang, H. 2018.
\newblock Language-driven synthesis of 3D scenes from scene databases.
\newblock \emph{ACM Transactions on Graphics (TOG)}, 37(6): 1--16.

\bibitem[{Ma et~al.(2023)Ma, Zhang, Sun, Ji, Wang, Jiang, Zhuang, and Ji}]{ma2023x}
Ma, Y.; Zhang, X.; Sun, X.; Ji, J.; Wang, H.; Jiang, G.; Zhuang, W.; and Ji, R. 2023.
\newblock X-mesh: Towards fast and accurate text-driven 3d stylization via dynamic textual guidance.
\newblock In \emph{Proceedings of the IEEE/CVF International Conference on Computer Vision}, 2749--2760.

\bibitem[{MeshyAI(2023)}]{Meshy}
MeshyAI. 2023.
\newblock Meshy.
\newblock \url{https://www.meshy.ai/}.

\bibitem[{Metzer et~al.(2023)Metzer, Richardson, Patashnik, Giryes, and Cohen-Or}]{metzer2023latent}
Metzer, G.; Richardson, E.; Patashnik, O.; Giryes, R.; and Cohen-Or, D. 2023.
\newblock Latent-nerf for shape-guided generation of 3d shapes and textures.
\newblock In \emph{Proceedings of the IEEE/CVF Conference on Computer Vision and Pattern Recognition}, 12663--12673.

\bibitem[{Mohammad~Khalid et~al.(2022)Mohammad~Khalid, Xie, Belilovsky, and Popa}]{mohammad2022clip}
Mohammad~Khalid, N.; Xie, T.; Belilovsky, E.; and Popa, T. 2022.
\newblock Clip-mesh: Generating textured meshes from text using pretrained image-text models.
\newblock In \emph{SIGGRAPH Asia 2022 conference papers}, 1--8.

\bibitem[{Nakajima(2024)}]{BabyAGI}
Nakajima, Y. 2024.
\newblock BabyAGI.
\newblock \url{https://github.com/yoheinakajima/babyagi}.

\bibitem[{OpenAI(2023)}]{openai2023vision}
OpenAI. 2023.
\newblock GPT-4 Vision.
\newblock \url{https://platform.openai.com/docs/guides/vision}.

\bibitem[{Oquab et~al.(2023)Oquab, Darcet, Moutakanni, Vo, Szafraniec, Khalidov, Fernandez, HAZIZA, Massa, El-Nouby et~al.}]{oquab2023dinov2}
Oquab, M.; Darcet, T.; Moutakanni, T.; Vo, H.~V.; Szafraniec, M.; Khalidov, V.; Fernandez, P.; HAZIZA, D.; Massa, F.; El-Nouby, A.; et~al. 2023.
\newblock DINOv2: Learning Robust Visual Features without Supervision.
\newblock \emph{Transactions on Machine Learning Research}.

\bibitem[{Osika(2024)}]{gpt-engineer}
Osika, A. 2024.
\newblock GPT-Engineer.
\newblock \url{https://github.com/gpt-engineer-org/gpt-engineer}.

\bibitem[{Paschalidou et~al.(2021)Paschalidou, Kar, Shugrina, Kreis, Geiger, and Fidler}]{paschalidou2021atiss}
Paschalidou, D.; Kar, A.; Shugrina, M.; Kreis, K.; Geiger, A.; and Fidler, S. 2021.
\newblock Atiss: Autoregressive transformers for indoor scene synthesis.
\newblock \emph{Advances in Neural Information Processing Systems}, 34: 12013--12026.

\bibitem[{Poole et~al.(2022)Poole, Jain, Barron, and Mildenhall}]{poole2022dreamfusion}
Poole, B.; Jain, A.; Barron, J.~T.; and Mildenhall, B. 2022.
\newblock DreamFusion: Text-to-3D using 2D Diffusion.
\newblock In \emph{The Eleventh International Conference on Learning Representations}.

\bibitem[{Qi et~al.(2018)Qi, Zhu, Huang, Jiang, and Zhu}]{qi2018human}
Qi, S.; Zhu, Y.; Huang, S.; Jiang, C.; and Zhu, S.-C. 2018.
\newblock Human-centric indoor scene synthesis using stochastic grammar.
\newblock In \emph{Proceedings of the IEEE Conference on Computer Vision and Pattern Recognition}, 5899--5908.

\bibitem[{Radford et~al.(2021)Radford, Kim, Hallacy, Ramesh, Goh, Agarwal, Sastry, Askell, Mishkin, Clark et~al.}]{radford2021learning}
Radford, A.; Kim, J.~W.; Hallacy, C.; Ramesh, A.; Goh, G.; Agarwal, S.; Sastry, G.; Askell, A.; Mishkin, P.; Clark, J.; et~al. 2021.
\newblock Learning transferable visual models from natural language supervision.
\newblock In \emph{International conference on machine learning}, 8748--8763. PMLR.

\bibitem[{Raj et~al.(2023)Raj, Kaza, Poole, Niemeyer, Ruiz, Mildenhall, Zada, Aberman, Rubinstein, Barron et~al.}]{raj2023dreambooth3d}
Raj, A.; Kaza, S.; Poole, B.; Niemeyer, M.; Ruiz, N.; Mildenhall, B.; Zada, S.; Aberman, K.; Rubinstein, M.; Barron, J.; et~al. 2023.
\newblock Dreambooth3d: Subject-driven text-to-3d generation.
\newblock In \emph{Proceedings of the IEEE/CVF International Conference on Computer Vision}, 2349--2359.

\bibitem[{Ritchie, Wang, and Lin(2019)}]{ritchie2019fast}
Ritchie, D.; Wang, K.; and Lin, Y.-a. 2019.
\newblock Fast and flexible indoor scene synthesis via deep convolutional generative models.
\newblock In \emph{Proceedings of the IEEE/CVF Conference on Computer Vision and Pattern Recognition}, 6182--6190.

\bibitem[{Shin et~al.(2020)Shin, Razeghi, Logan~IV, Wallace, and Singh}]{shin2020autoprompt}
Shin, T.; Razeghi, Y.; Logan~IV, R.~L.; Wallace, E.; and Singh, S. 2020.
\newblock AutoPrompt: Eliciting Knowledge from Language Models with Automatically Generated Prompts.
\newblock In \emph{Proceedings of the 2020 Conference on Empirical Methods in Natural Language Processing (EMNLP)}, 4222--4235.

\bibitem[{Shinn et~al.(2024)Shinn, Cassano, Gopinath, Narasimhan, and Yao}]{shinn2024reflexion}
Shinn, N.; Cassano, F.; Gopinath, A.; Narasimhan, K.; and Yao, S. 2024.
\newblock Reflexion: Language agents with verbal reinforcement learning.
\newblock \emph{Advances in Neural Information Processing Systems}, 36.

\bibitem[{Shtedritski, Rupprecht, and Vedaldi(2023)}]{shtedritski2023does}
Shtedritski, A.; Rupprecht, C.; and Vedaldi, A. 2023.
\newblock What does clip know about a red circle? visual prompt engineering for vlms.
\newblock In \emph{Proceedings of the IEEE/CVF International Conference on Computer Vision}, 11987--11997.

\bibitem[{Su et~al.(2022)Su, Kasai, Wu, Shi, Wang, Xin, Zhang, Ostendorf, Zettlemoyer, Smith et~al.}]{su2022selective}
Su, H.; Kasai, J.; Wu, C.~H.; Shi, W.; Wang, T.; Xin, J.; Zhang, R.; Ostendorf, M.; Zettlemoyer, L.; Smith, N.~A.; et~al. 2022.
\newblock Selective annotation makes language models better few-shot learners.
\newblock \emph{arXiv preprint arXiv:2209.01975}.

\bibitem[{SudoAI(2023)}]{Sudo}
SudoAI. 2023.
\newblock Audo AI.
\newblock \url{https://www.sudo.ai/}.

\bibitem[{Tang et~al.(2024)Tang, Chen, Chen, Wang, Zeng, and Liu}]{tang2024lgm}
Tang, J.; Chen, Z.; Chen, X.; Wang, T.; Zeng, G.; and Liu, Z. 2024.
\newblock LGM: Large Multi-View Gaussian Model for High-Resolution 3D Content Creation.
\newblock \emph{arXiv preprint arXiv:2402.05054}.

\bibitem[{Tang et~al.(2023{\natexlab{a}})Tang, Nie, Markhasin, Dai, Thies, and Nie{\ss}ner}]{tang2023diffuscene}
Tang, J.; Nie, Y.; Markhasin, L.; Dai, A.; Thies, J.; and Nie{\ss}ner, M. 2023{\natexlab{a}}.
\newblock Diffuscene: Scene graph denoising diffusion probabilistic model for generative indoor scene synthesis.
\newblock \emph{arXiv preprint arXiv:2303.14207}.

\bibitem[{Tang et~al.(2023{\natexlab{b}})Tang, Ren, Zhou, Liu, and Zeng}]{tang2023dreamgaussian}
Tang, J.; Ren, J.; Zhou, H.; Liu, Z.; and Zeng, G. 2023{\natexlab{b}}.
\newblock Dreamgaussian: Generative gaussian splatting for efficient 3d content creation.
\newblock \emph{arXiv preprint arXiv:2309.16653}.

\bibitem[{TripoAI(2024)}]{tripo3d}
TripoAI. 2024.
\newblock Tripo3d.
\newblock \url{https://www.tripo3d.ai/}.

\bibitem[{Wang et~al.(2022{\natexlab{a}})Wang, Chai, He, Chen, and Liao}]{wang2022clip}
Wang, C.; Chai, M.; He, M.; Chen, D.; and Liao, J. 2022{\natexlab{a}}.
\newblock Clip-nerf: Text-and-image driven manipulation of neural radiance fields.
\newblock In \emph{Proceedings of the IEEE/CVF Conference on Computer Vision and Pattern Recognition}, 3835--3844.

\bibitem[{Wang et~al.(2023)Wang, Jiang, Chai, He, Chen, and Liao}]{wang2023nerf}
Wang, C.; Jiang, R.; Chai, M.; He, M.; Chen, D.; and Liao, J. 2023.
\newblock Nerf-art: Text-driven neural radiance fields stylization.
\newblock \emph{IEEE Transactions on Visualization and Computer Graphics}.

\bibitem[{Wang et~al.(2019)Wang, Lin, Weissmann, Savva, Chang, and Ritchie}]{wang2019planit}
Wang, K.; Lin, Y.-A.; Weissmann, B.; Savva, M.; Chang, A.~X.; and Ritchie, D. 2019.
\newblock Planit: Planning and instantiating indoor scenes with relation graph and spatial prior networks.
\newblock \emph{ACM Transactions on Graphics (TOG)}, 38(4): 1--15.

\bibitem[{Wang et~al.(2018)Wang, Savva, Chang, and Ritchie}]{wang2018deep}
Wang, K.; Savva, M.; Chang, A.~X.; and Ritchie, D. 2018.
\newblock Deep convolutional priors for indoor scene synthesis.
\newblock \emph{ACM Transactions on Graphics (TOG)}, 37(4): 1--14.

\bibitem[{Wang et~al.(2022{\natexlab{b}})Wang, Wei, Schuurmans, Le, Chi, Narang, Chowdhery, and Zhou}]{wang2022self}
Wang, X.; Wei, J.; Schuurmans, D.; Le, Q.~V.; Chi, E.~H.; Narang, S.; Chowdhery, A.; and Zhou, D. 2022{\natexlab{b}}.
\newblock Self-Consistency Improves Chain of Thought Reasoning in Language Models.
\newblock In \emph{The Eleventh International Conference on Learning Representations}.

\bibitem[{Wang, Yeshwanth, and Nie{\ss}ner(2021)}]{wang2021sceneformer}
Wang, X.; Yeshwanth, C.; and Nie{\ss}ner, M. 2021.
\newblock Sceneformer: Indoor scene generation with transformers.
\newblock In \emph{2021 International Conference on 3D Vision (3DV)}, 106--115. IEEE.

\bibitem[{Wang et~al.(2024)Wang, Lu, Wang, Bao, Li, Su, and Zhu}]{wang2024prolificdreamer}
Wang, Z.; Lu, C.; Wang, Y.; Bao, F.; Li, C.; Su, H.; and Zhu, J. 2024.
\newblock Prolificdreamer: High-fidelity and diverse text-to-3d generation with variational score distillation.
\newblock \emph{Advances in Neural Information Processing Systems}, 36.

\bibitem[{Wen et~al.(2023)Wen, Liu, Sridhar, and Fu}]{wen2023anyhome}
Wen, Z.; Liu, Z.; Sridhar, S.; and Fu, R. 2023.
\newblock AnyHome: Open-Vocabulary Generation of Structured and Textured 3D Homes.
\newblock \emph{arXiv preprint arXiv:2312.06644}.

\bibitem[{Wu et~al.(2023)Wu, Bansal, Zhang, Wu, Zhang, Zhu, Li, Jiang, Zhang, and Wang}]{wu2023autogen}
Wu, Q.; Bansal, G.; Zhang, J.; Wu, Y.; Zhang, S.; Zhu, E.; Li, B.; Jiang, L.; Zhang, X.; and Wang, C. 2023.
\newblock Autogen: Enabling next-gen llm applications via multi-agent conversation framework.
\newblock \emph{arXiv preprint arXiv:2308.08155}.

\bibitem[{Xi et~al.(2023)Xi, Chen, Guo, He, Ding, Hong, Zhang, Wang, Jin, Zhou et~al.}]{xi2023rise}
Xi, Z.; Chen, W.; Guo, X.; He, W.; Ding, Y.; Hong, B.; Zhang, M.; Wang, J.; Jin, S.; Zhou, E.; et~al. 2023.
\newblock The rise and potential of large language model based agents: A survey.
\newblock \emph{arXiv preprint arXiv:2309.07864}.

\bibitem[{Yang et~al.(2023{\natexlab{a}})Yang, Zhang, Li, Zou, Li, and Gao}]{yang2023set}
Yang, J.; Zhang, H.; Li, F.; Zou, X.; Li, C.; and Gao, J. 2023{\natexlab{a}}.
\newblock Set-of-mark prompting unleashes extraordinary visual grounding in gpt-4v.
\newblock \emph{arXiv preprint arXiv:2310.11441}.

\bibitem[{Yang et~al.(2024)Yang, Wang, Li, Wang, and Yang}]{yang2024fine}
Yang, L.; Wang, Y.; Li, X.; Wang, X.; and Yang, J. 2024.
\newblock Fine-grained visual prompting.
\newblock \emph{Advances in Neural Information Processing Systems}, 36.

\bibitem[{Yang et~al.(2022)Yang, Gan, Wang, Hu, Lu, Liu, and Wang}]{yang2022empirical}
Yang, Z.; Gan, Z.; Wang, J.; Hu, X.; Lu, Y.; Liu, Z.; and Wang, L. 2022.
\newblock An empirical study of gpt-3 for few-shot knowledge-based vqa.
\newblock In \emph{Proceedings of the AAAI Conference on Artificial Intelligence}, volume~36, 3081--3089.

\bibitem[{Yang et~al.(2023{\natexlab{b}})Yang, Li, Lin, Wang, Lin, Liu, and Wang}]{yang2023dawn}
Yang, Z.; Li, L.; Lin, K.; Wang, J.; Lin, C.-C.; Liu, Z.; and Wang, L. 2023{\natexlab{b}}.
\newblock The dawn of lmms: Preliminary explorations with gpt-4v (ision).
\newblock \emph{arXiv preprint arXiv:2309.17421}, 9(1): 1.

\bibitem[{Yao et~al.(2023)Yao, Zhao, Yu, Du, Shafran, Narasimhan, and Cao}]{yao2023react}
Yao, S.; Zhao, J.; Yu, D.; Du, N.; Shafran, I.; Narasimhan, K.; and Cao, Y. 2023.
\newblock ReAct: Synergizing Reasoning and Acting in Language Models.
\newblock In \emph{International Conference on Learning Representations (ICLR)}.

\bibitem[{Yeh et~al.(2012)Yeh, Yang, Watson, Goodman, and Hanrahan}]{yeh2012synthesizing}
Yeh, Y.-T.; Yang, L.; Watson, M.; Goodman, N.~D.; and Hanrahan, P. 2012.
\newblock Synthesizing open worlds with constraints using locally annealed reversible jump mcmc.
\newblock \emph{ACM Transactions on Graphics (TOG)}, 31(4): 1--11.

\bibitem[{Yi et~al.(2023)Yi, Fang, Wu, Xie, Zhang, Liu, Tian, and Wang}]{yi2023gaussiandreamer}
Yi, T.; Fang, J.; Wu, G.; Xie, L.; Zhang, X.; Liu, W.; Tian, Q.; and Wang, X. 2023.
\newblock Gaussiandreamer: Fast generation from text to 3d gaussian splatting with point cloud priors.
\newblock \emph{arXiv preprint arXiv:2310.08529}.

\bibitem[{Zhai et~al.(2024)Zhai, {\"O}rnek, Wu, Di, Tombari, Navab, and Busam}]{zhai2024commonscenes}
Zhai, G.; {\"O}rnek, E.~P.; Wu, S.-C.; Di, Y.; Tombari, F.; Navab, N.; and Busam, B. 2024.
\newblock Commonscenes: Generating commonsense 3d indoor scenes with scene graphs.
\newblock \emph{Advances in Neural Information Processing Systems}, 36.

\bibitem[{Zhan et~al.(2024)Zhan, Dai, Ye, Zhou, Zhang, Liu, Zhang, Yuan, Zhang, Li et~al.}]{zhan2024anygpt}
Zhan, J.; Dai, J.; Ye, J.; Zhou, Y.; Zhang, D.; Liu, Z.; Zhang, X.; Yuan, R.; Zhang, G.; Li, L.; et~al. 2024.
\newblock AnyGPT: Unified Multimodal LLM with Discrete Sequence Modeling.
\newblock \emph{arXiv preprint arXiv:2402.12226}.

\bibitem[{Zhang et~al.(2020)Zhang, Yang, Ma, Luo, Huth, Vouga, and Huang}]{zhang2020deep}
Zhang, Z.; Yang, Z.; Ma, C.; Luo, L.; Huth, A.; Vouga, E.; and Huang, Q. 2020.
\newblock Deep generative modeling for scene synthesis via hybrid representations.
\newblock \emph{ACM Transactions on Graphics (TOG)}, 39(2): 1--21.

\bibitem[{Zhang et~al.(2022)Zhang, Zhang, Li, and Smola}]{zhang2022automatic}
Zhang, Z.; Zhang, A.; Li, M.; and Smola, A. 2022.
\newblock Automatic Chain of Thought Prompting in Large Language Models.
\newblock In \emph{The Eleventh International Conference on Learning Representations}.

\bibitem[{Zhao et~al.(2024)Zhao, Huang, Xu, Lin, Liu, and Huang}]{zhao2024expel}
Zhao, A.; Huang, D.; Xu, Q.; Lin, M.; Liu, Y.-J.; and Huang, G. 2024.
\newblock Expel: Llm agents are experiential learners.
\newblock In \emph{Proceedings of the AAAI Conference on Artificial Intelligence}, volume~38, 19632--19642.

\bibitem[{Zhou et~al.(2022)Zhou, Muresanu, Han, Paster, Pitis, Chan, and Ba}]{zhou2022large}
Zhou, Y.; Muresanu, A.~I.; Han, Z.; Paster, K.; Pitis, S.; Chan, H.; and Ba, J. 2022.
\newblock Large Language Models are Human-Level Prompt Engineers.
\newblock In \emph{The Eleventh International Conference on Learning Representations}.

\bibitem[{Zhou, While, and Kalogerakis(2019)}]{zhou2019scenegraphnet}
Zhou, Y.; While, Z.; and Kalogerakis, E. 2019.
\newblock Scenegraphnet: Neural message passing for 3d indoor scene augmentation.
\newblock In \emph{Proceedings of the IEEE/CVF International Conference on Computer Vision}, 7384--7392.

\bibitem[{Zhu et~al.(2023)Zhu, Chen, Shen, Li, and Elhoseiny}]{zhu2023minigpt}
Zhu, D.; Chen, J.; Shen, X.; Li, X.; and Elhoseiny, M. 2023.
\newblock MiniGPT-4: Enhancing Vision-Language Understanding with Advanced Large Language Models.
\newblock In \emph{The Twelfth International Conference on Learning Representations}.

\end{thebibliography}

\clearpage

\appendix



We provide supplementary material for \textit{Chat2Layout}, covering a summary of related furniture layout generation works,
implementation details, prompts for floor placement, and pose alignment. 
Additionally, we provide a user study and discuss our user interface.
Lastly, we discuss the limitations of our method.

\section{Summary of Layout Generation Works}
\label{sec:ac}

\begin{table*}[htp]
\centering
\renewcommand{\arraystretch}{1.2}
\tabcolsep=0.05cm
\scalebox{0.67}{
\begin{tabular}{cccccc}
\toprule
 & Layout Completion & \begin{tabular}[c]{@{}c@{}}(Partial) Layout\\ Re-arrangement\end{tabular} & Text Instruction & 3D Placement & \begin{tabular}[c]{@{}c@{}}Open-Set Placement/\\ Multi-Conventional Interaction\end{tabular} \\ \hline
\begin{tabular}[c]{@{}c@{}}\cite{yeh2012synthesizing}, Grains \cite{li2019grains}, \cite{qi2018human}, \\ \cite{luo2020end}, CommonScenes \cite{zhai2024commonscenes}\end{tabular} &  &  &  &  & \\ \hline
\begin{tabular}[c]{@{}c@{}}
\cite{zhang2020deep}, \cite{ritchie2019fast},\\ 
Planit \cite{wang2019planit}, SceneGraphNet \cite{zhou2019scenegraphnet}
\end{tabular} & \Checkmark &  &  &  & \\ \hline
\begin{tabular}[c]{@{}c@{}}\cite{fu2017adaptive}, \cite{wang2018deep},\\ 
\cite{dhamo2021graph}, Atiss \cite{paschalidou2021atiss}\end{tabular}
& \Checkmark & \Checkmark &  &  & \\ \hline
SceneFormer \cite{wang2021sceneformer}, \cite{ma2016action}, \cite{ma2018language} & \Checkmark &  & \Checkmark &  & \\ \hline
\cite{chang2014learning}, \cite{chang2015text} &  &  & \Checkmark &  & \\ \hline
\begin{tabular}[c]{@{}c@{}}InstructScene \cite{lin2023instructscene}, DiffuScene \cite{tang2023diffuscene},\\ 
LayoutGPT \cite{feng2024layoutgpt}\end{tabular} & \Checkmark & \Checkmark & \Checkmark &  & \\ \hline
AnyHome \cite{wen2023anyhome} & \Checkmark & \Checkmark & \Checkmark & \Checkmark & \\
\hline
Chat2Layout & \Checkmark & \Checkmark & \Checkmark & \Checkmark & \Checkmark \\ 
\bottomrule
\end{tabular}}
\vspace{5pt}
\caption{
\textbf{Summary of furniture layout generation methods classified by supported applications:} layout completion, layout rearrangement, text instruction, 3D placement, open-set placement, and multi-conventional interaction. 
\name supports all applications, with particular strengths in open-set object placement and multi-conventional user interaction,
a feature unsupported by other methods.
}
\label{tab:summary}
\end{table*}

Table \ref{tab:summary} provides a summary of optimized-based methods \cite{qi2018human,ma2016action,chang2015text,chang2014learning,yeh2012synthesizing,fisher2012example,luo2020end,wang2019planit,ma2018language,fu2017adaptive}, learning-based methods \cite{zhang2020deep,wang2018deep,ritchie2019fast,zhou2019scenegraphnet,li2019grains,dhamo2021graph,paschalidou2021atiss,wang2021sceneformer,zhai2024commonscenes,tang2023diffuscene,lin2023instructscene}, and LLM-based methods  \cite{feng2024layoutgpt,wen2023anyhome} on furniture layout generation.
In comparison, \name supports a wider range of applications than these baselines.
Specifically, \name supports layout completion, layout rearrangement, text instruction, 3D placement (onto or under others, or on walls), open-set placement, and multi-conventional interaction.
Notably, supporting open-set placement means that \name is not restricted to predefined furniture objects nor reliant on retrieving items from a 3D furniture dataset. Moreover, our agent system facilitates continuous user interaction through multi-run conversations, allowing users to iteratively refine the layout plan until desired outcome is reached. 
It should be noted that \textit{AnyHome} supports only single-turn conversations for layout adjustments without memorization, whereas our method supports multi-turn interactions through the agent, favoring long-term and complex interactions.

\section{Implementation Details}
\label{sec:details}

We employ GPT-4 vision \cite{openai2023vision} as the MLLM to implement our agent. We develop the pipeline in Python and implement user-interface/visualization in Open3D. The system is deployed on Windows/Linux workstations without GPUs.
To ensure the creation of satisfactory and style-consistent 3D meshes from a text prompts, we generate $16$ candidates using `\textit{Text-2-3D\textless text\textgreater }' and have the agent select the most suitable one.
We also incorporate style descriptions (if applicable) into `\textit{Text-2-3D\textless text\textgreater }' to maintain stylistic coherence. For instance, in a Gothic-style living room, if a user requests adding a chair, `\textit{\textless text\textgreater }' in `\textit{Text-2-3D\textless text\textgreater }' would specify `\textit{a Gothic-style chair}' rather than simply `\textit{a chair}'.
For support set searching, we compile databases containing $50$ to $100$ candidate examples per question. In Equation 1, we set $\alpha=0.3$ and $\beta=0.2$ to prioritize textual similarity while integrating textual and visual environmental information. In addition, we utilize the `clip-vit-base-patch16' model to calculate CLIP similarity. 
In our O2O-Search algorithm, the example is categorized by characteristics. For instance, `0’ indicates a removal task and `1’ an abstract addition task, and we have 8 categories in total for task decomposition.

\section{Pose Alignment}
\label{sec:pose}

\begin{figure}[H]
\centering
\includegraphics[width=1.0\linewidth]{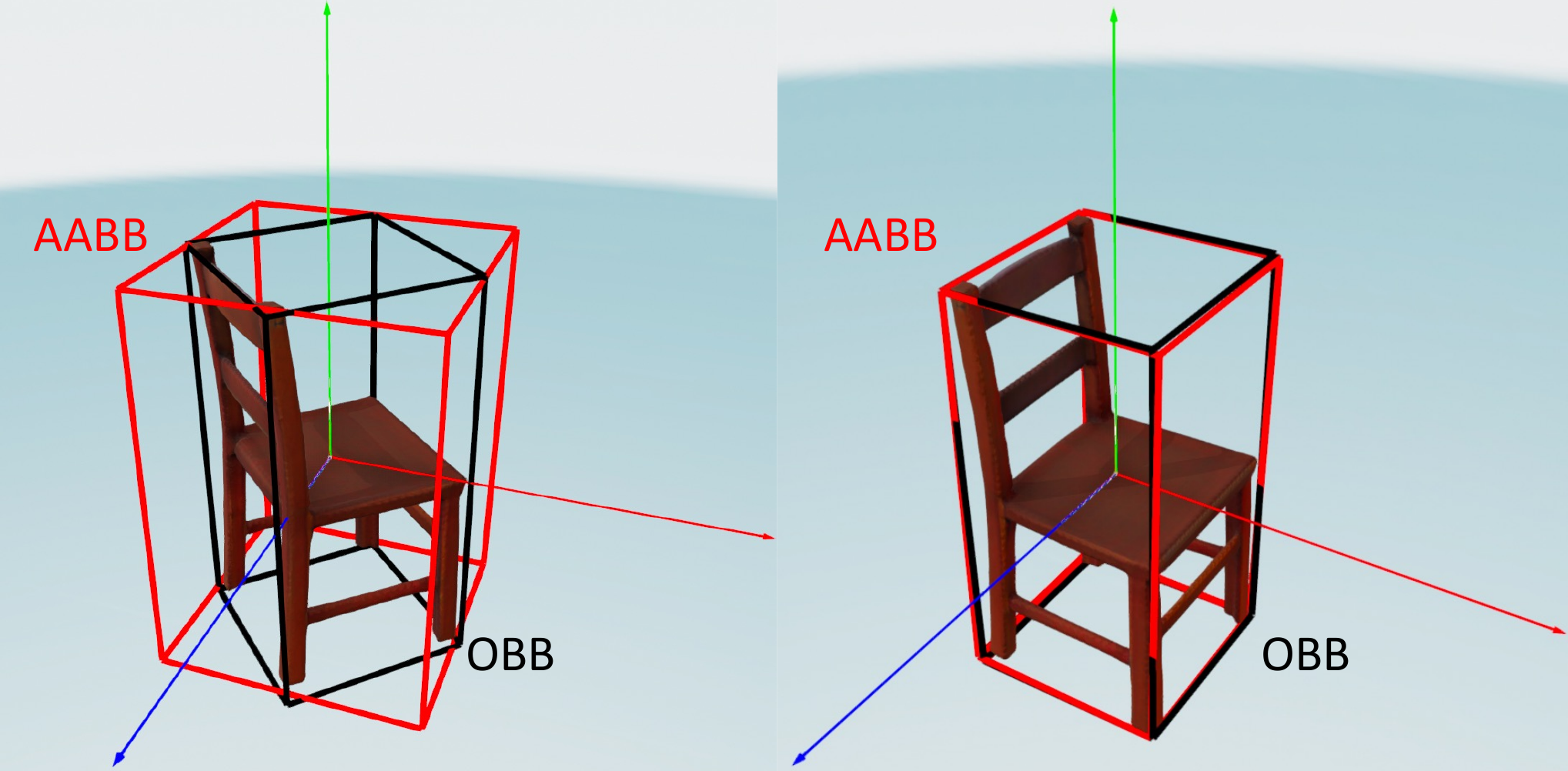}
\begin{small}
\begin{tabular}{p{0.33\linewidth}
<{\centering}p{0.12\linewidth}
<{\centering}p{0.33\linewidth}<{\centering} }
Before && After
\end{tabular}
\end{small}
\caption{\textbf{Visualization of object pose alignment.} We show a chair before (left) and after (right) pose alignment. Initially, we compute the Axis-Aligned Bounding Box (AABB) and the Optimal/Oriented Bounding Box (OBB) \cite{fabri2009cgal}. By analying the difference between the AABB and OBB, we identify the necessary rotation to align the chair's orientation. After applying the rotation, the chair's AABB and OBB become nearly identical, both aligned with the axis.}
\label{fig:pose}
\end{figure}

Before arranging the layout, we generate a set of required furniture items using `\textit{Text-2-3D\textless text\textgreater }'. However, the initial poses of these objects may be arbitrary. To address this, we 
perform pose alignment, as illustrated in Figure \ref{fig:pose}.
We begin by calculating both the Axis-Aligned Bounding Box (AABB) and the Optimal/Oriented Bounding Box (OBB) \cite{fabri2009cgal} for each object. By comparing the AABB and OBB, we determine the pose difference between them. Then, we rotate the object based on the calculated pose difference, effectively aligning the AABB and OBB to achieve a more suitable initial pose for the object within the scene.

\section{Prompts for Floor Placement}
\label{sec:prompt}

\begin{figure*}
\centering 
\includegraphics[width=\textwidth]{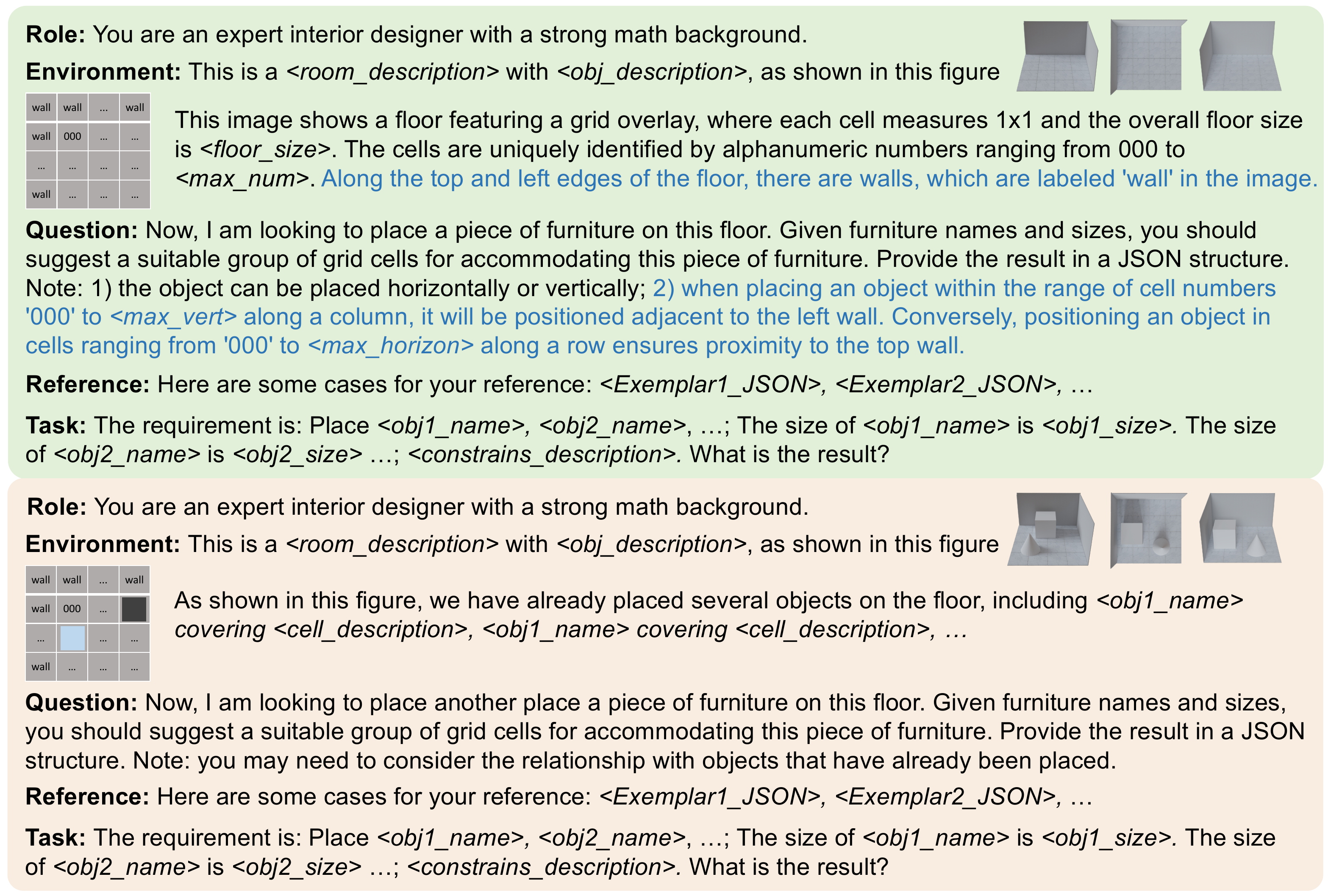}
\caption{\textbf{Vision-question prompting of floor placement.} Here we show two exemplar prompts for placing an new furniture item on an empty or occupied floor, respectively. The words highlighted in blue will be omitted when there is no wall present.}
\label{fig:vqp}
\end{figure*}

Figure \ref{fig:vqp} illustrates the exemplar prompts used for object placement on the floor. This vision-text prompting approach, employed by our floor placement algorithm, adheres to the vision-question paradigm, consisting of a role, environment information, visual-text prompts, and a reference set provided by our O2O-Search. All other prompts in our system follow a similar structure, but with variations in text descriptions and visual captures specific to their respective tasks.
Leveraging the memory capabilities of LLMs, our agent enables users to ask follow-up questions such as ``Provide more results" or request layout completion, as demonstrated in the bottom part of Figure \ref{fig:vqp}.

\section{User Interface}
\label{sec:ui}

\begin{figure}[t]
\centering
\includegraphics[width=0.96\linewidth]{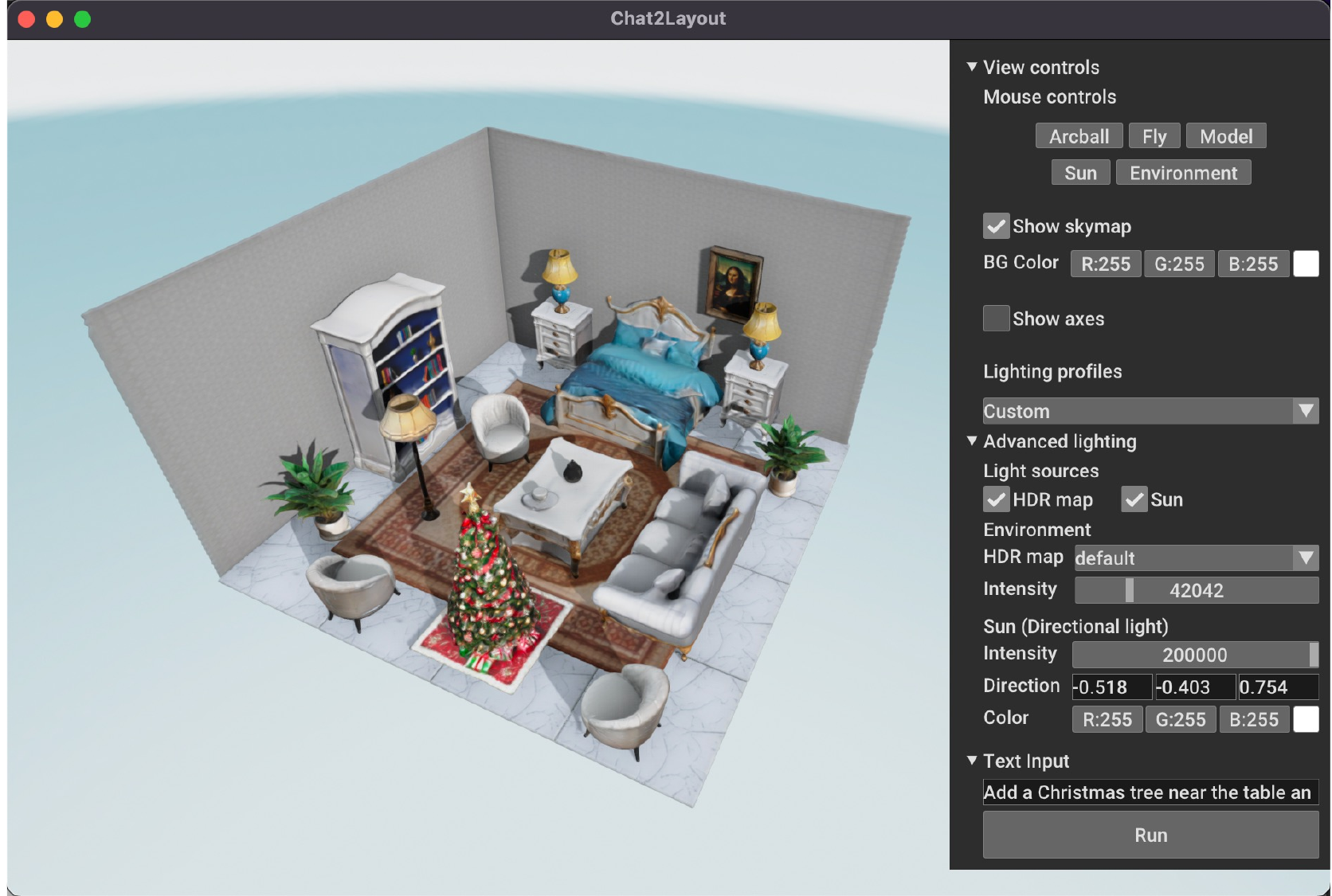}
\caption{\textbf{Our user interface.} Our agent system facilitates user interactions through intuitive text instructions. The interface also empowers users to adjust the viewpoint and lighting within the 3D environment for a better visualization.}
\label{fig:ui}
\end{figure}

We design a user-friendly interface using Open3D, as shown in Figure \ref{fig:ui}. Users can easily input text instructions and initiate an interaction with our agent system by clicking the button. The generated visual results are displayed in the left widget, providing a feedback on the layout design. Our interface also offers controls for adjusting the viewpoint and lighting, allowing users to optimize the visualization for better understanding of the 3D layout. For a more comprehensive demonstration of our interface, please refer to our accompanying video demo.

\section{User Study}
\label{sec:us}

\begin{figure}[H]
\centering 
\includegraphics[width=1.0\linewidth]{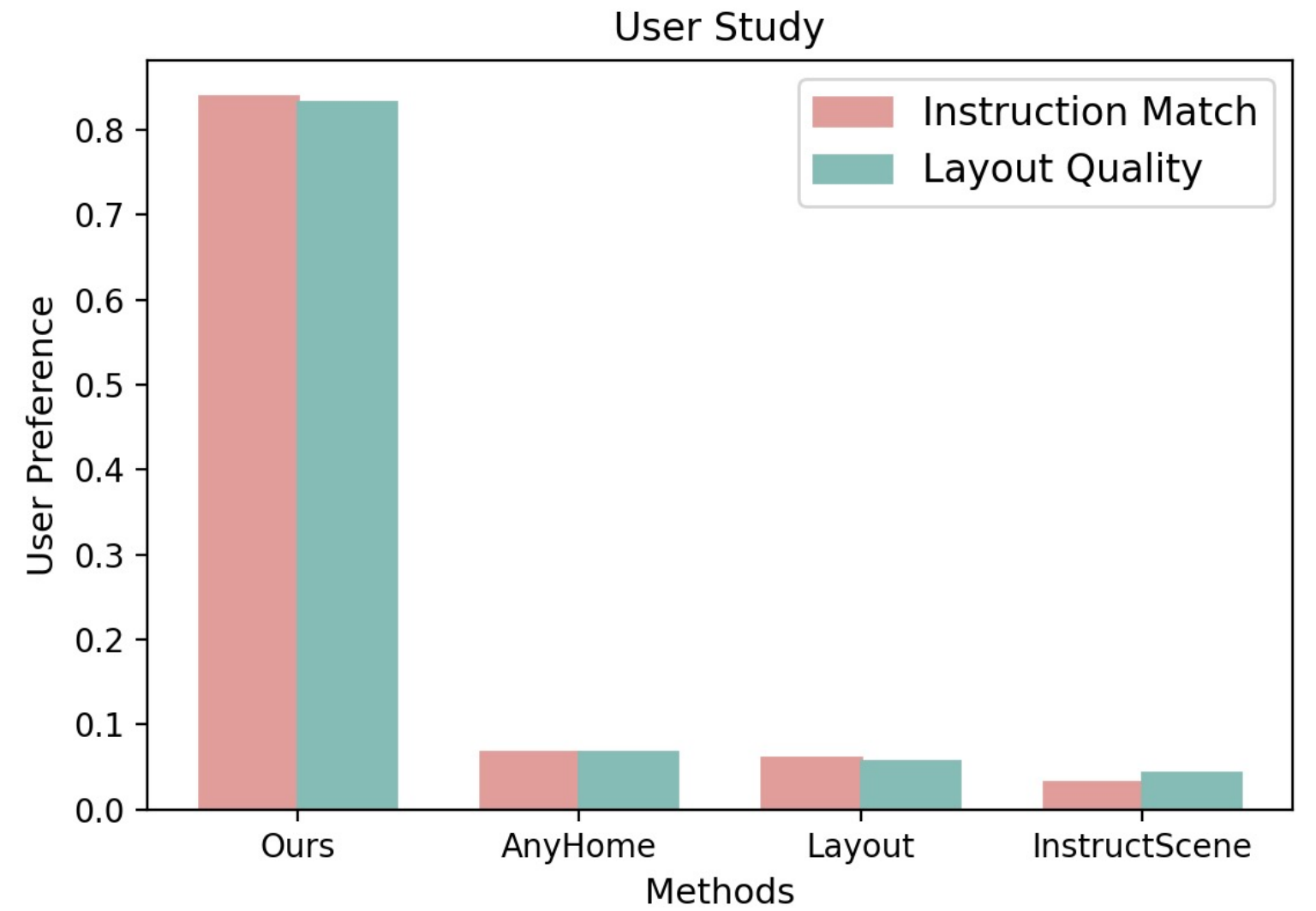}
\caption{\textbf{User study for comparison.} We conduct a user study comparing \textit{AnyHome}, \textit{LayoutGPT}, and \textit{InstructScene} with \textit{Chat2Layout} on layout generation quality. Participants are presented with various layout options and asked to evaluate: 1) which layout plan best aligns with user instructions; and 2) which layout is the most reasonable and realistic. Our approach outperforms the other baselines, garnering higher user preference in both questions.}
\label{fig:us}
\end{figure}

\begin{table}[H]
\centering
\tabcolsep=0.05cm
\scalebox{0.75}{
\begin{tabular}{c|ccccc}
\hline
\textbf{Metric} & \begin{tabular}[c]{@{}c@{}}\textbf{Mental} \\  \textbf{Demand↓}\end{tabular} &  \begin{tabular}[c]{@{}c@{}}\textbf{Physical} \\  \textbf{Demand↓}\end{tabular} & \begin{tabular}[c]{@{}c@{}}\textbf{Temporal} \\  \textbf{Demand↓}\end{tabular} & \textbf{Performance↑} & \textbf{Frustration↓} \\ \hline
\textbf{Mean Score} & 1.73 & 1.91 & 3.64 & 4.09 & 2.18 \\ \hline
\end{tabular}}
\caption{\textbf{NASA-TLX mean scores.} We conduct a user study for the user interaction assessment. Participants are required to complete NASA-TLX questionnaires (1=very low,5=very high) after using our user interface.}
\label{table:nasa_tlx_mean_scores}
\end{table}

We conduct a user study, illustrated in Figure \ref{fig:us}, comparing \textit{AnyHome}, \textit{LayoutGPT}, and \textit{InstructScene} with \textit{Chat2Layout}. Participants are presented with $20$ questions, each accompanied by text instructions and four layout options. They are asked to select the option that best: 1) matches the given instructions; and 2) presents a reasonable and realistic layout plan, free from unreasonable collisions or boundary exceedances. After collecting $41$ valid responses, we analyze the user preference rates for both instruction matching and layout quality. According to the results, our approach demonstrate superior performance, surpassing the other methods in user preference across both criteria.

We have conducted another user study to evaluate the interaction in Table \ref{table:nasa_tlx_mean_scores}. We invited six non-professional participants to create scenes using our system. Initially, we provided a 10-minute tutorial. Participants were then asked to create 5 predefined scenes and 5 scenes of their choice. Each participant then completed a questionnaire using the five-point NASA Task Load Index (NASA-TLX, 1=very low to 5=very high) \cite{hart2006nasa} to evaluate usability and perceived workload. 
In Table 2, the mental demand was low, indicating that users need little effort when using our system. While our system is not perfect in terms of efficiency due to the unstable GPT-4 API, it performs satisfactorily with a high performance value and demonstrates satisfying user satisfaction with a low frustration value.

\section{Limitations}
\label{sec:limit}

\begin{figure}[H]
\centering 
\includegraphics[width=1.0\linewidth]{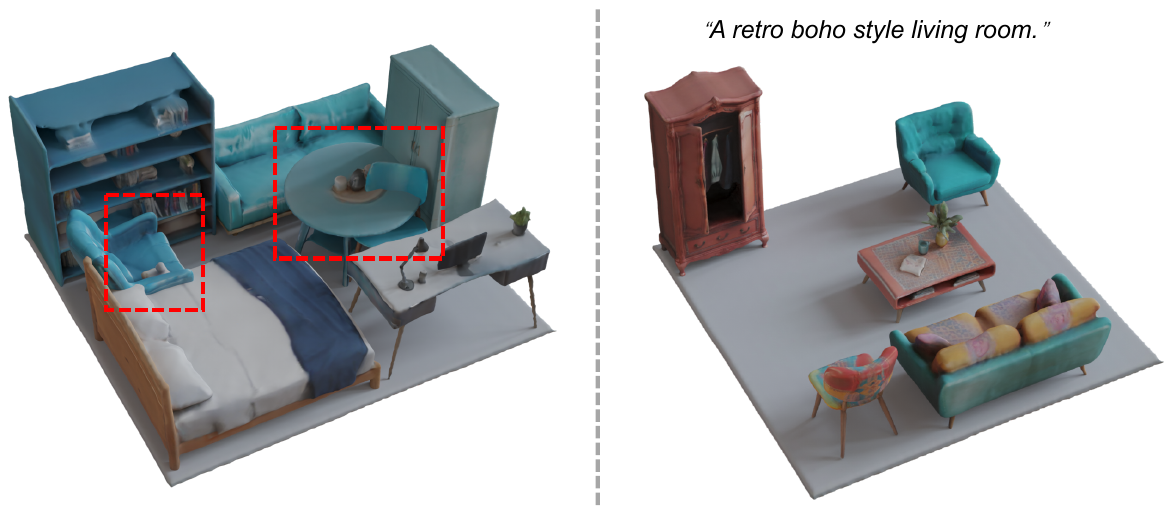}
\caption{\textbf{Limitations.} Our method faces challenges in certain scenarios. First, when dealing with a large number of furniture items in a severely limited space, collisions can occur, as highlighted by the red rectangle in the left figure. Second, current text-to-3D generation methods struggle to produce a cohesive set of furniture items that adhere to a consistent style based on textual descriptions. The wardrobe in the right figure exemplifies this issue, which appears visually inconsistent with the other items in the scene.}
\label{fig:limit}
\end{figure}

While our agent system facilitates a wider range of user interactions and generates more reasonable layout plans than existing methods, it still faces several challenges, as shown in Figure \ref{fig:limit}.

Firstly, when tasked with arranging a large number of furniture pieces within a severely limited space, collision issue becomes evident. This red rectangle in the left figure highlights how spatial constraints can lead to suboptimal furniture placement and overlapping elements. This occurs because when the available space is insufficient for the size and quantity of objects, it poses a challenge for our agent system to devise a reasonable layout plan. In future iterations, implementing a checking mechanism and user reminders will be crucial to effectively address this issue.

Secondly, our system's reliance on existing text-to-3D generation methods presents another challenge. Currently, these methods struggle to produce a harmonized set of furniture items that adhere to a consistent style when given a series of text instructions. While we have partially mitigated this issue by generating multiple candidate items and allowing the agent to select the most suitable one, the problem is not entirely resolved, as occasionally one or more items may not fit seamlessly within the scene.
For example, as shown in Figure \ref{fig:limit}-right, the generated wardrobe lacks stylistic consistency with the other items.
It would be a valuable advancement to develop a text-to-3D method capable of producing a cohesive set of style-conscious assets, which we intend to explore in the future.

Finally, efficiency remains a limitation of our system. Generating a scene from scratch currently takes around one minute due to the interaction time with the GPT-4 vision API and the generation of a set of candidate 3D furniture—the exact time depends on the complexity of the generated scene. The instability of the GPT-4 vision API sometimes induces errors due to unstable network and negatively impacts user experience. We believe that efficiency can be significantly improved in the future with the development of more efficient large language models (LLMs).

\end{document}